\documentclass[runningheads]{llncs}

\usepackage{eccv}

\usepackage{eccvabbrv}

\usepackage{graphicx}
\usepackage{booktabs}

\usepackage[accsupp]{axessibility}  

\usepackage{hyperref}

\usepackage{orcidlink}
\usepackage{bm}
\usepackage{amsmath}

\usepackage[normalem]{ulem} 

\providecommand{\realnum}{\ensuremath{\mathbb{R}}}

\providecommand{\bmcal}[1]{\ensuremath{\bm{\mathcal{#1}}}}

\newcommand{\MethodName}{MIGS\xspace}

\begin{document}

\title{MIGS: Multi-Identity Gaussian Splatting via Tensor Decomposition}


\author{Aggelina Chatziagapi\inst{1} \and
Grigorios G. Chrysos\inst{2} \and
Dimitris Samaras\inst{1} \\
}


\authorrunning{A.~Chatziagapi et al.}


\institute{Stony Brook University 
\email{\{aggelina,samaras\}@cs.stonybrook.edu} 
\and University of Wisconsin-Madison \email{chrysos@wisc.edu}}

\maketitle

\begin{abstract}
\vspace{-10pt}
We introduce \MethodName (Multi-Identity Gaussian Splatting), a novel method that learns a single neural representation for multiple identities, using only monocular videos. Recent 3D Gaussian Splatting (3DGS) approaches for human avatars require per-identity optimization. However, learning a multi-identity representation presents advantages in robustly animating humans under arbitrary poses. We propose to construct a high-order tensor that combines all the learnable 3DGS parameters for all the training identities. By assuming a low-rank structure and factorizing the tensor, we model the complex rigid and non-rigid deformations of multiple subjects in a unified network, significantly reducing the total number of parameters. Our proposed approach leverages information from all the training identities and enables robust animation under challenging unseen poses, outperforming existing approaches. It can also be extended to learn unseen identities.
Project page: \url{https://aggelinacha.github.io/MIGS/}.


\end{abstract}

\section{Introduction}

Synthesizing and animating digital humans has broad applications, ranging from immersive telepresence in AR/VR to video games and the movie industry. However, this task presents significant challenges, since humans are highly deformable entities in the real world. Capturing their large variety of spatio-temporal (4D) rigid and non-rigid deformations is a widely studied topic in computer vision and graphics~\cite{SMPL:2015, AMASS:ICCV:2019, weng2022humannerf}. A vast number of works~\cite{SMPL:2015,SMPL-X:2019,loper2023smpl,MANO:SIGGRAPHASIA:2017,AMASS:ICCV:2019,STAR:2020} have proposed to learn parametric models of the human body, like SMPL~\cite{SMPL:2015}, using large datasets of human scans. These models
can be used as a prior, conditioning neural representations of human avatars~\cite{weng2022humannerf,yu2023monohuman,mu2023actorsnerf,qian20233dgs}.

\begin{figure}[tb]
  \centering
      \includegraphics[width=\linewidth]{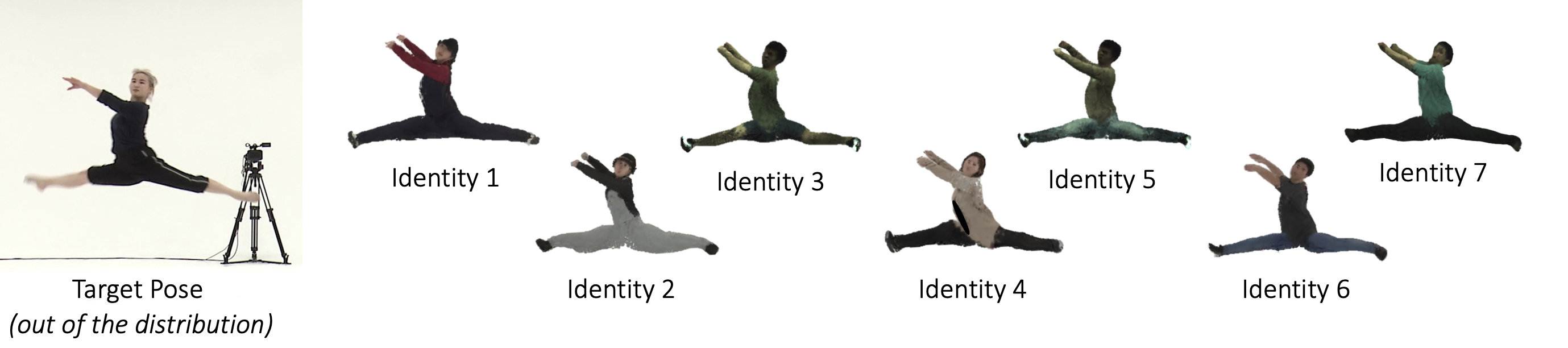}
        \vspace{-10pt}
  \caption{We introduce \MethodName (multi-identity Gaussian splatting) that learns a \emph{single} neural representation for \emph{multiple identities} based on tensor decomposition. \MethodName enables robust animation of human avatars under novel poses, out of the training distribution.}
  \label{fig:teaser}
  \vspace{-15pt}
\end{figure}


Advances in implicit representations and neural rendering, including Neural Radiance Fields (NeRFs)~\cite{mildenhall2020nerf}, have enabled photo-realistic modeling of the appearance and geometry of humans~\cite{nerface,nerfies,zielonka2023insta,weng2022humannerf,yu2023monohuman,mu2023actorsnerf}.
The non-rigid and rigid deformations of the human body are represented by one or more multi-layer perceptrons (MLPs). The learned dynamic representation can be used for novel view synthesis and animation under novel poses~\cite{weng2022humannerf,mu2023actorsnerf,yu2023monohuman}. More recently, 3D Gaussian Splatting (3DGS)~\cite{qian20233dgs} achieves real-time rendering of radiance fields with a high quality. By representing a scene with a set of 3D Gaussians and proposing a fast rasterization method, 3DGS significantly outperforms NeRFs that require expensive optimization and long inference times. Recent works extend 3DGS to human avatars~\cite{qian20233dgs,hu2023gaussianavatar,hu2023gauhuman,Zielonka2023Drivable3D}, by conditioning on a human articulation prior~\cite{SMPL:2015}.
However, given a monocular setting, these approaches can only learn single-identity representations. Only a few works have proposed more generic neural representations, but require multiple views for training~\cite{kwon2021neural,raj2021pixel,zheng2023gpsgaussian}.

In this work, we address the less studied problem of learning a single neural representation of multiple identities from monocular videos.
In multi-view settings, related methods usually learn blending mechanisms, extract features from nearby views, and infer a novel view in feed-forward manner~\cite{wang2021ibrnet,chen2021mvsnerf,zheng2023gpsgaussian,kwon2021neural}. Constrained by the input views, such multi-view approaches cannot animate humans under novel poses. In contrast, in a monocular setting, we cannot leverage nearby views as input.
Learning a single representation of multiple subjects is an interesting research topic,
since we need to preserve individual characteristics in a generic model,
and presents the following advantages. (a) By leveraging information from multiple identities, it learns a better representation of the human body articulation, enabling robust animation under unseen, even out-of-distribution, poses.
(b) Even though recent neural rendering methods have become faster, they still require per-identity optimization~\cite{qian20233dgs,hu2023gauhuman,weng2022humannerf}, which can lead to a prohibitive number of learnable parameters for a large number of identities.


We introduce \MethodName (Multi-Identity Gaussian Splatting), a novel method that learns a \emph{multi-identity} 3DGS representation of human avatars. To the best of our knowledge, this is the first method that extends 3DGS to multiple identities, learned in a monocular setting. Inspired by ideas that go back to TensorFaces~\cite{vasilescu2002multilinear}, we construct a high-order tensor of all the learnable parameters of our 3DGS representation. The core premise of our method is to learn a tensor decomposition~\cite{kolda2009tensor,panagakis2021tensor} of the high-order tensor.
The intuition is that different identities share common characteristics. Thus, we assume a \emph{low-rank structure} of the high-order tensor and decompose it into a sum of rank-one tensors~\cite{hitchcock1927tensor_polyadic, kolda2009tensor}.
These rank-one tensors correspond to the only learnable parameters of our optimization procedure. In this way, we learn a \emph{single} unified representation of multiple subjects, and use only a \emph{fraction} of the total parameters that would be required by training multiple per-identity 3DGS representations. By leveraging information from multiple identities, \MethodName enables robust animation under challenging out-of-distribution poses.
We also demonstrate how it can be extended to learn novel identities.

In brief, our contributions are as follows:
\begin{itemize}
    \item We propose to construct a high-order tensor that combines all the learnable parameters of a 3DGS representation for multiple human avatars. 
 

    \item By factorizing this tensor with a tensor decomposition, we learn a single unified representation for multiple identities, and significantly reduce the total number of learnable parameters compared to single-identity representations.
   
    \item By leveraging information from multiple identities, our proposed \MethodName enables robust animation under novel poses, outperforming existing approaches.  
\end{itemize}
\section{Related Work}

\noindent
\textbf{Human Body Modeling.}
Reconstructing and animating humans has been a long standing problem in computer vision and graphics. Earlier approaches develop stereo techniques, relying on multi-view studio setups~\cite{kanade1997constructing,matusik2000image,carranza2003free}. They model the non-rigid and rigid deformations of humans with mesh~\cite{vlasic2008articulated} or volume~\cite{de2008performance} representations. Motion retargeting has also been explored~\cite{10.1145/280814.280820,10.1145/311535.311539}. With the advent of generative adversarial networks~\cite{goodfellow2014generative}, many approaches propose image-to-image translation methods~\cite{pix2pix2017,wang2018video,wang2018high} for 2D motion transfer~\cite{chan2019everybody,Prokudin_2021_WACV,neverova2018dense}, based on pose detection systems~\cite{8765346,simon2017hand,cao2017realtime,wei2016cpm}. A vast number of works~\cite{SMPL:2015,SMPL-X:2019,loper2023smpl,MANO:SIGGRAPHASIA:2017,AMASS:ICCV:2019,STAR:2020} have proposed to learn parametric models of the human body, like SMPL~\cite{SMPL:2015}, using large datasets of 3D human scans. These models can represent a human with a set of parameters for the body shape, joint position and rotation. They have been beneficial
for reconstructing and tracking humans over time in videos~\cite{Bogo:ECCV:2016,hmrKanazawa17,rajasegaran2022tracking}. They have also been used as a human articulation prior to condition implicit neural representations of human avatars~\cite{weng2022humannerf,yu2023monohuman,mu2023actorsnerf,qian20233dgs}.

\noindent
\textbf{Neural Rendering of Human Avatars.} Recent advances in implicit representations and neural rendering have enabled photo-realistic modeling of the appearance and geometry of humans. Neural Radiance Fields (NeRFs)~\cite{mildenhall2020nerf} have been first proposed for static scenes, showing remarkable results for novel view synthesis~\cite{mildenhall2020nerf,barron2021mipnerf,barron2022mipnerf360,lindell2022bacon}. They have been extended to dynamic scenes~\cite{pumarola2021d,li2021neural,li2022neural}, making them a popular choice for modeling human bodies~\cite{peng2021neural,xu2021h,2021narf,liu2021neuralactor,kania2022conerf,peng2023implicit,mpsnerf,jayasundara2023flexnerf} and faces~\cite{nerface,nerfies,park2021hypernerf}. While many of those NeRF-based approaches use multi-view videos, recent works attempt to capture the 4D dynamics and appearance of humans from monocular videos~\cite{nerface,weng2022humannerf,yu2023monohuman,mu2023actorsnerf,jiang2022neuman,su2021nerf,wang2022neural,adnerf}. Most of these works align the observed poses of different video frames in a single canonical space, and learn a conditional 3D representation. The non-rigid and rigid deformations of the human body are represented by one or more multi-layer perceptrons (MLPs). The learned dynamic representation can be used for novel view synthesis and animation under novel poses~\cite{weng2022humannerf,mu2023actorsnerf,yu2023monohuman}. However, all these approaches require expensive per-identity optimization. A few works have proposed more generic representations~\cite{wang2021ibrnet,chen2021mvsnerf,trevithick2021grf,yu2021pixelnerf,kwon2021neural,mu2023actorsnerf,kim2023yoto}, but these require static settings or multiple views for training. In contrast, we propose to learn a single representation for multiple identities, from just monocular videos of challenging poses.

\noindent
\textbf{3D Gaussian Splatting.}
More recently, several approaches propose techniques for faster training and rendering of radiance fields, with 3D Gaussian Splatting (3DGS) becoming very popular~\cite{qian20233dgs}. 3DGS achieves real-time rendering of complex static scenes with a high visual quality. It represents a scene with a set of 3D Gaussians. Each Gaussian is associated with learnable parameters, namely a position (mean), an anisotropic covariance matrix, an opacity, and a color. The 3D Gaussians are projected to 2D splats in the image space through a fast rasterization procedure. Recent works extend 3DGS to dynamic human avatars~\cite{qian20233dgs,hu2023gaussianavatar,hu2023gauhuman,Zielonka2023Drivable3D}, by conditioning the 3DGS representation on a human articulation prior~\cite{SMPL:2015}. However, these approaches only learn single-identity representations. Learning a more generic neural representation would require multiple views for training~\cite{kwon2021neural,raj2021pixel,zheng2023gpsgaussian}. In this work, we extend 3DGS to multiple identities, in a monocular setting, demonstrating the benefits of such a multi-identity representation in robustly animating human avatars under novel poses.

\noindent
\textbf{Tensor Structures.} Tensors are multi-dimensional arrays that can naturally represent data of multiple dimensions~\cite{kolda2009tensor,panagakis2021tensor}. They have been used in related fields to represent visual data and capture higher-order similarities or interactions~\cite{georgopoulos2020multilinear,chrysos2021conditional}. TensorFaces~\cite{vasilescu2002multilinear} is an early approach that approximates different modes of variation using a multilinear tensor decomposition.
Tensor-based operations have been proposed within deep neural networks to leverage the topological structure in the data~\cite{kossaifi2017tensor,kossaifi2020tensor}. Tensor decomposition extends the concept of matrix decomposition~\cite{turk1991eigenfaces} to higher-order tensors for dimensionality reduction and data compression.
It has been applied to neural networks to
significantly reduce the number of learnable parameters and mitigate the curse of dimensionality~\cite{novikov2015tensorizing,lebedev2014speeding,astrid2017cp}. A widely used tensor decomposition is the CANDECOMP/PARAFAC (CP) decomposition~\cite{hitchcock1927tensor_polyadic,carroll1970analysis,harshman1970foundations,kolda2009tensor} that decomposes a tensor into a sum of rank-one tensors. Computing the CP decomposition of a tensor requires optimization algorithms, like Alternating Least Square (ALS)~\cite{mohlenkamp2013musings}. In this work, we propose to construct a high-order tensor of all the learnable parameters of a 3DGS representation, and learn a CP tensor decomposition to model multiple identities in a unified network and use a reduced number of parameters. To the best of our knowledge, this is the first approach that applies such a decomposition to 3DGS and learns a multi-identity 3DGS representation.


\section{Preliminary}\label{sec:prelim}

\subsubsection{Notation.} We follow the notation introduced by Kolda and Bader~\cite{kolda2009tensor}. We denote scalars as $w \in \realnum$, vectors as $\bm{w} \in \realnum ^{I_1}$, and matrices as $\bm{W} \in \realnum ^{I_1 \times I_2}$. Tensors are symbolized by calligraphic letters, \eg  $\bmcal{W} \in \realnum ^{I_1 \times I_2 \times \dots \times I_d}$. Tensor fibers $\bm{w}_{i,j,:}$ are higher-order analogues of matrix rows and columns. Tensor slices $\bm{W}_{i,:,:}$ are 2-dimensional sections of a tensor.

\subsection{3D Gaussian Splatting}\label{sec:prelim_3dgs}

3D Gaussian Splatting (3DGS)~\cite{kerbl20233d} has been first proposed to represent static scenes, achieving fast and high-quality novel view synthesis. A scene is represented by a set of anisotropic 3D Gaussians. Each Gaussian is defined by a position (mean) $\bm{\mu}$ and a covariance matrix $\bm{\Sigma}$:
\begin{equation}
    G(\bm{x}) = e ^ {(\bm{x} - \bm{\mu})^T \bm{\Sigma} ^{-1} (\bm{x} - \bm{\mu})}\;,
\end{equation}
and associated with an opacity $\alpha$ and a color $\bm{c}$ that are used for $\alpha$-blending. The 3D Gaussians are projected to 2D splats in the image space following EWA splatting~\cite{zwicker2001ewa}, where Zicker \etal demonstrate how the corresponding 2D covariance matrix can be derived from $\bm{\Sigma}$. The whole representation is differentiable, optimized via gradient descent. In order to ensure that $\bm{\Sigma}$ has a physical meaning, it is constrained to be positive semi-definite, by defining a parametric ellipsoid with a diagonal scaling matrix $\bm{S}$ and a rotation matrix $\bm{R}$:
\begin{equation}
    \bm{\Sigma} = \bm{R} \bm{S} \bm{S}^T \bm{R}^T \;.
\end{equation}
In practice, the parameters of each Gaussian are stored as a position vector $\bm{\mu} \in \realnum^3$, a 3D scaling vector $\bm{s} \in \realnum^3$ and a quaternion $\bm{q} \in \realnum^4$ to represent the rotation~\cite{kerbl20233d}. The locations of the Gaussians are initialized by a sparse point cloud, obtained by Structure-from-Motion~\cite{schonberger2016structure}. Then, they are optimized based on successive iterations of rendering, guided by a photometric loss and using an adaptive densification scheme~\cite{kerbl20233d}. The color $C$ of a pixel is computed by blending all Gaussians $g$ that overlap the pixel, after sorting them by depth: 
\begin{equation}
    C = \sum_{g} \bm{c}_{g} \alpha_{g} \prod_{j=1}^{g-1} (1 - \alpha_{j}) \;,
\end{equation}
where $\bm{c}_{g}$ is the color and $\alpha_{g}$ is the learned opacity weighted by the probability density of the $g$-th projected 2D Gaussian splat. The color $c_g$ can be a view-dependent color computed from learned spherical harmonic (SH) coefficients~\cite{fridovich2022plenoxels,muller2022instant,kerbl20233d} or can be a learned RGB color vector~\cite{luiten2023dynamic,qian20233dgs}.

\subsection{Deformable 3D Gaussian Splatting for Humans}\label{sec:prelim_3dgs_humans}

Many recent works extend 3DGS to represent human avatars~\cite{qian20233dgs,hu2023gaussianavatar,hu2023gauhuman,Zielonka2023Drivable3D}, combining the 3D Gaussians with a human articulation prior. They follow a widely adopted paradigm, mapping the observation to the canonical space and learning a representation of the human body in the canonical pose~\cite{weng2022humannerf,wang2022arah,jiang2022neuman,weng2020vid2actor}. They use linear blend skinning (LBS) to deform the human body under arbitrary poses, based on the SMPL model~\cite{loper2023smpl}. 

\begin{figure}[tb]
  \centering
  \includegraphics[width=\linewidth]{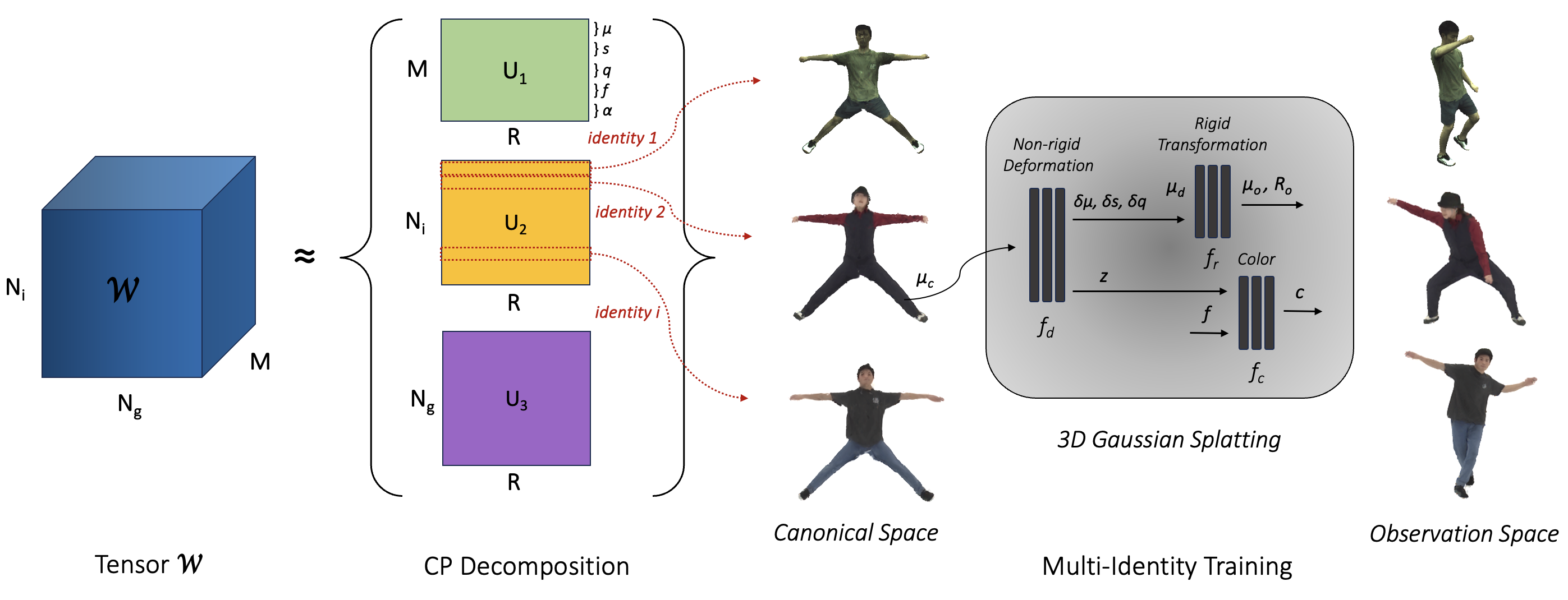}
  \caption{\textbf{Overview of \MethodName.} Given monocular videos of multiple identities, we learn a unified 3DGS representation for human avatars based on CP tensor decomposition. We construct a tensor $\bm{\mathcal{W}}  \in \realnum ^ {N_i \times N_g \times M}$, where $N_i$ is the number of identities, $N_g$ the number of 3D Gaussians and $M$ the number of parameters per Gaussian. In practice, we assume a low-rank structure of the tensor $\bm{\mathcal{W}}$ and thus, we only learn the matrices $\bm{U}_1 \in \realnum ^{M \times R}$, $\bm{U}_2 \in \realnum ^{N_i \times R}$, $\bm{U}_3 \in \realnum ^{N_g \times R}$ that approximate $\bm{\mathcal{W}}$ through the CP decomposition with $R << N_g$. By leveraging information from the diverse deformations of different subjects, \MethodName enables robust animation under novel challenging poses.
  }
  \label{fig:overview}
\vspace{-10pt}
\end{figure}

Given a monocular video of a human, corresponding SMPL parameters and foreground masks per frame, we learn a set of 3D Gaussians $\{G^{(g)}\}_{g=1}^{N_g}$ that represent the human body in the canonical pose. We initialize them by randomly sampling $N_g$ points on the SMPL mesh surface under canonical pose. Following HumanNeRF~\cite{weng2022humannerf} and 3DGS-Avatar~\cite{qian20233dgs}, we decompose the complex human deformation to rigid and non-rigid parts, which model the skeletal motion and non-rigid cloth deformation correspondingly (see overview in~\cref{fig:overview} right).

\noindent
\textbf{Non-rigid Deformation.} First, we learn a non-rigid deformation module $f_d$ that maps the set of 3D Gaussians under canonical pose $\{G_{c}\}$ to a non-rigidly deformed space $\{G_{d}\}$. More specifically, given the parameters of a Gaussian in the canonical space (position $\bm{\mu}_c$, scale $\bm{s}_c$, and rotation quartenion $\bm{q}_{c}$) and a latent code $\bm{z}_p$, $f_d$ outputs offsets and a feature vector $\bm{z}$ as follows:
\begin{equation}
    (\delta \bm{\mu}, \delta \bm{s}, \delta \bm{q}, \bm{z}) = f_d(\bm{\mu}_c; \bm{z}_p)\;,
\end{equation}
where $\bm{\mu}_d = \bm{\mu}_c + \delta \bm{\mu}$, $\bm{s}_d = \bm{s}_c \cdot e^{\delta \bm{s}}$, $\bm{q}_d = \bm{q}_c \cdot [1; \delta \bm{q} ] $, and $\bm{z}_p$ is a latent code that encodes the SMPL parameters for pose and shape as the output of an hierarchical encoder~\cite{LEAP:CVPR:21,qian20233dgs}.

\noindent
\textbf{Rigid Transformation.} The non-rigidly deformed Gaussians $\{G_{d}\}$ are mapped to the observation space via a rigid transformation module. The rigid transformation is based on LBS~\cite{loper2023smpl}, where $B$ rigid bone transformations $\{\bm{B}_b\}_{b=1}^{B}$ are linearly blended via a set of skinning weights learned by a neural skinning field represented as an MLP $f_r$~\cite{chen2021snarf,weng2022humannerf,qian20233dgs}:
\begin{equation}
   \bm{T} = \sum_{b=1}^{B} f_r(\bm{\mu}_d)_b \bm{B}_b\;.
\end{equation}
Then, the position and rotation of the Gaussians in the observation space are computed as: $\bm{\mu}_o = \bm{T} \bm{\mu}_d$ and $\bm{R}_o = \bm{T}_{:3,:3} \bm{R}_d$, where $\bm{R}_d$ is the rotation matrix derived from the quartenion $\bm{q}_d$.

\noindent
\textbf{Color.} Following 3DGS-Avatar~\cite{qian20233dgs}, we predict the color $\bm{c}$ of a Gaussian using an MLP $f_c$, conditioned on a per-Gaussian learnable feature vector $\bm{f} \in \realnum^{32}$, the output $\bm{z}$ of the non-rigid network $f_d$, and the SH basis $\bm{\gamma}(\hat{\bm{d}})$:
\begin{equation}\label{eq:color_mlp}
   \bm{c} = f_c(\bm{f}, \bm{z}, \bm{\gamma}(\hat{\bm{d}})) \;,
\end{equation}
where $\bm{d}$ is the viewing direction wrt the camera center and $\hat{\bm{d}} = T_{:3,:3}^{-1} \bm{d}$ is the canonicalized viewing direction to ensure consistency under the applied transformations~\cite{qian20233dgs}. Note that in contrast to 3DGS-Avatar, we do not learn any per-frame latent code, in order to avoid any overfitting to the training frames. Our goal is to learn a more general representation of the human body, given multiple identities.

\section{Method}

We present \MethodName, a novel method that learns a multi-identity 3DGS representation for human avatars, based on tensor decomposition. An overview of our approach is illustrated in~\cref{fig:overview}. Given only monocular videos of different subjects, we learn a unified neural representation that models their appearance and articulation. Inspired by ideas that go back to TensorFaces~\cite{vasilescu2002multilinear}, we construct a high-order tensor of the Gaussian parameters of all our training identities. Given that different identities share common characteristics, we assume a low-rank structure of the high-order tensor. Applying a tensor decomposition~\cite{kolda2009tensor,panagakis2021tensor}, we decompose it into a sum of rank-one tensors. In this way, we share weights between different identities, using a minimum number of total learnable parameters for our Gaussian splatting. By learning a unified representation, \MethodName enables robust animation of the training subjects under completely unseen poses. We also demonstrate how \MethodName can be extended to novel identities.

\subsection{Multi-Identity Representation}

Previous works~\cite{qian20233dgs,hu2023gaussianavatar,hu2023gauhuman,Zielonka2023Drivable3D} propose single-identity 3DGS representations for human avatars, where the parameters of the 3D Gaussians are learned from a monocular video of a single subject. \Cref{sec:prelim_3dgs_humans} defines the 3DGS representation for a single identity. In this section, we demonstrate how we construct our \emph{multi-identity} representation based on high-order tensors. Tensors are a multi-dimensional generalization of matrices~\cite{kolda2009tensor,panagakis2021tensor} that can represent data of multiple dimensions and capture high-order similarities or interactions. 

For each identity $i$, we assume a set of 3D Gaussians $\{G^{(i, g)}\}_{g=1}^{N_g}$ that model the human avatar under canonical pose. Each Gaussian $G^{(i, g)}$ is defined by the following learnable parameters: a position (mean) $\bm{\mu}^{(i, g)} \in \realnum^3$, a scaling vector $\bm{s}^{(i, g)} \in \realnum^3$, a quartenion $\bm{q}^{(i, g)} \in \realnum^4$, a feature vector $\bm{f}^{(i, g)} \in \realnum^{32}$ and an opacity $\alpha^{(i, g)} \in \realnum$. The color $\bm{c}^{(i, g)}$ of the Gaussian is given by the output of a color MLP as in~\cref{eq:color_mlp}. Given videos of $N_i$ identities, we construct a tensor:
\begin{equation}\label{eq:tensor_w}
   \bm{\mathcal{W}}  \in \realnum ^ {N_i \times N_g \times M}, \text{where } \bm{w}_{i, g, :} = \left[ \bm{\mu}^{(i, g)}; \bm{s}^{(i, g)}; \bm{q}^{(i, g)}; \bm{f}^{(i, g)}; \alpha^{(i, g)} \right] \;,
\end{equation}
where $M$ derives by adding the dimensions of the parameters for a Gaussian $g$ and an identity $i$: $M = 3 + 3 + 4 + 32 + 1 = 43$.

\subsection{Tensor Decomposition}\label{sec:tensor_decomp}

Increasing the number of identities leads to a prohibitive number of learnable parameters in the tensor $\bm{\mathcal{W}}$. We assume a low-rank structure of $\bm{\mathcal{W}}$ and apply CANDECOMP/PARAFAC (CP) tensor decomposition~\cite{hitchcock1927tensor_polyadic, kolda2009tensor} to decompose it into a sum of a finite number of rank-one tensors. More specifically, we first matricize the tensor $\bm{\mathcal{W}}$ using the unfolding in the second dimension, such that $\bm{W}_{(2)} \in \realnum ^ {N_g \times (N_i M)}$. Then, we set:
\begin{equation}\label{eq:cp_decomp}
   \bm{W}_{(2)} \approx \bm{U}_3 \left( \bm{U}_2 \odot \bm{U}_1 \right)^T \;,
\end{equation}
where $\odot$ denotes the Khatri-Rao product~\cite{khatri1968solutions}, $\bm{U}_1 \in \realnum ^{M \times R}$, $\bm{U}_2 \in \realnum ^{N_i \times R}$, $\bm{U}_3 \in \realnum ^{N_g \times R}$ are the factor matrices, and $R$ is a positive integer that corresponds to the selected rank of the decomposition.
In an exact CP decomposition, where there is equality in~\cref{eq:cp_decomp}, the smallest number of rank-one tensors that can generate $\bmcal{W}$ as their sum is equal with the rank of the tensor $\bmcal{W}$~\cite{kolda2009tensor}.

In practice, we only learn the matrices $\bm{U}_1, \bm{U}_2, \bm{U}_3$. Thus, we learn $(M+N_i+N_g) R$ 
parameters, instead of $M N_i N_g$ parameters, which leads to a significant decrease for a large number of identities $N_i$, when $R << N_g$. In our experiments, $R=100$ and $N_g=5 \times 10^4$. For $N_i = 30$ identities, MIGS learns only $(M+N_i+N_g) R \approx 5 \times 10 ^ 6 $ parameters, compared to $M N_i N_g \approx 6.5 \times 10^7$ that would be required by single-identity 3DGS representations, leading to a decrease \emph{by at least one order of magnitude} in the total number of learnable parameters.
During each iteration, we use \cref{eq:cp_decomp} to derive $\bm{W}_{(2)}$ and then rewrite it as a tensor $\bm{\mathcal{W}}$ (see \cref{eq:tensor_w}). Each fiber $\bm{w}_{i,g,:}$ corresponds to the parameters of a 3D Gaussian $G^{(i, g)}$ for an identity $i = 1, \dots, N_i$ and $g = 1, \dots, N_g$. 

\subsection{Optimization}

\noindent
\textbf{Initialization.} We initialize $\bm{U}_1, \bm{U}_2, \bm{U}_3$ by using the parameters of only one identity. More specifically, for the identity $i = 1$, we randomly sample $N_g$ points on the SMPL mesh surface under canonical pose to initialize the locations $\bm{\mu}^{(1, g)}$ for $g = 1, \dots, N_g$. We follow~\cite{kerbl20233d} to initialize the rest of the parameters $\bm{s}^{(1, g)}$, $\bm{q}^{(1, g)}$, $\bm{f}^{(1, g)}$, $\alpha^{(1, g)}$ and we construct the tensor slice $\bm{W}_{1,:,:}$. We then compute the CP decomposition of $\bm{W}_{1,:,:}$ by applying Robust Tensor Power Iteration (CPPower) using TensorLy~\cite{kossaifi2016tensorly}~\footnote{\url{https://tensorly.org/stable/modules/generated/tensorly.decomposition.CPPower.html}}. This gives us the initial matrices $\bm{U}_1$, $\bm{U}_3$, and the first row of $\bm{U}_2$ that we populate to all the rows to get our initial $\bm{U}_2$. We absorb the output weights of CPPower in $\bm{U}_3$.

\noindent
\textbf{Training.}
We follow the 3DGS implementation for human avatars as described in \cref{sec:prelim_3dgs_humans}. We jointly optimize $\bm{U}_1, \bm{U}_2, \bm{U}_3$, as well as the parameters of $f_d$, $f_r$, and $f_c$, by iteratively rendering frames from each training identity. Given the estimated $\bm{U}_1, \bm{U}_2, \bm{U}_3$ at each iteration, we derive the parameters of the 3D Gaussians for a specific identity, as described in \cref{sec:tensor_decomp}. We use the same loss function as 3DGS-Avatar~\cite{qian20233dgs} that consists of a photometric RGB loss, a mask loss, a skinning weight regularization, and an as-isometric-as-possible regularization loss (see supplementary document for more implementation details).

\noindent
\textbf{Per-identity Updates.}
Our proposed approach provides the option to update the parameters for only a specific identity $i$ at a given iteration, by using only the row $i$ of $\bm{U}_2$ in \cref{eq:cp_decomp}. In our experiments, we observe similar results in both cases, \ie using either the full $\bm{U}_2$ at each iteration or the corresponding row of $\bm{U}_2$ of the current identity $i$. However, we note that individual per-identity updates can be useful in other cases.

\subsection{Personalization}\label{sec:pers_cont}


Trained on multiple identities simultaneously, \MethodName learns a large variety of rigid and non-rigid deformations. To enhance the output visual quality for a particular subject, we can further fine-tune our color MLP $f_c$ (see~\cref{eq:color_mlp}), keeping the rest of the parameters frozen, using a short video of that subject. We observe that this ``personalization'' procedure is useful when we train our network for a large number of identities. Increasing the number of identities leads to an increase in robustness under novel poses, but a smoother result, where our generic network might not capture high-frequency details. This is expected, since for a given number of parameters, we learn a single unified representation of a large number of identities with diverse characteristics. We address this with the personalization that, with only a few iterations, successfully captures identity-specific details.

\noindent
\textbf{Novel Identity.} Learning a novel identity (that is not part of the initial training set) is also possible, by just adding an additional row to $\bm{U}_2$. Given a short video of the new identity, we optimize only the corresponding row of $\bm{U}_2$ and the color MLP $f_c$, keeping the rest of our representation frozen. In this way, we learn the characteristics of the new identity, without forgetting the large variety of the human body deformations captured by our multi-identity representation.

\begin{figure}[tb]
  \centering
  \includegraphics[width=\linewidth]{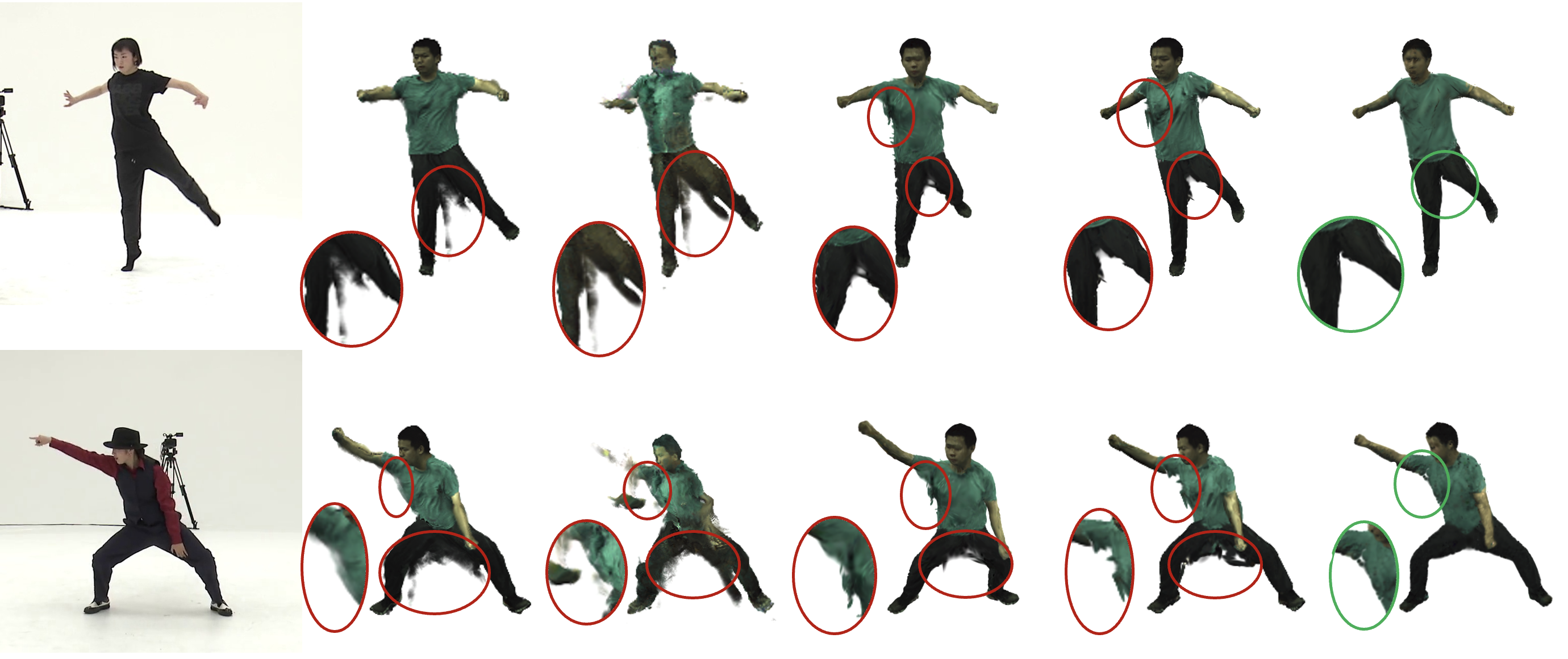}
\includegraphics[width=\linewidth]{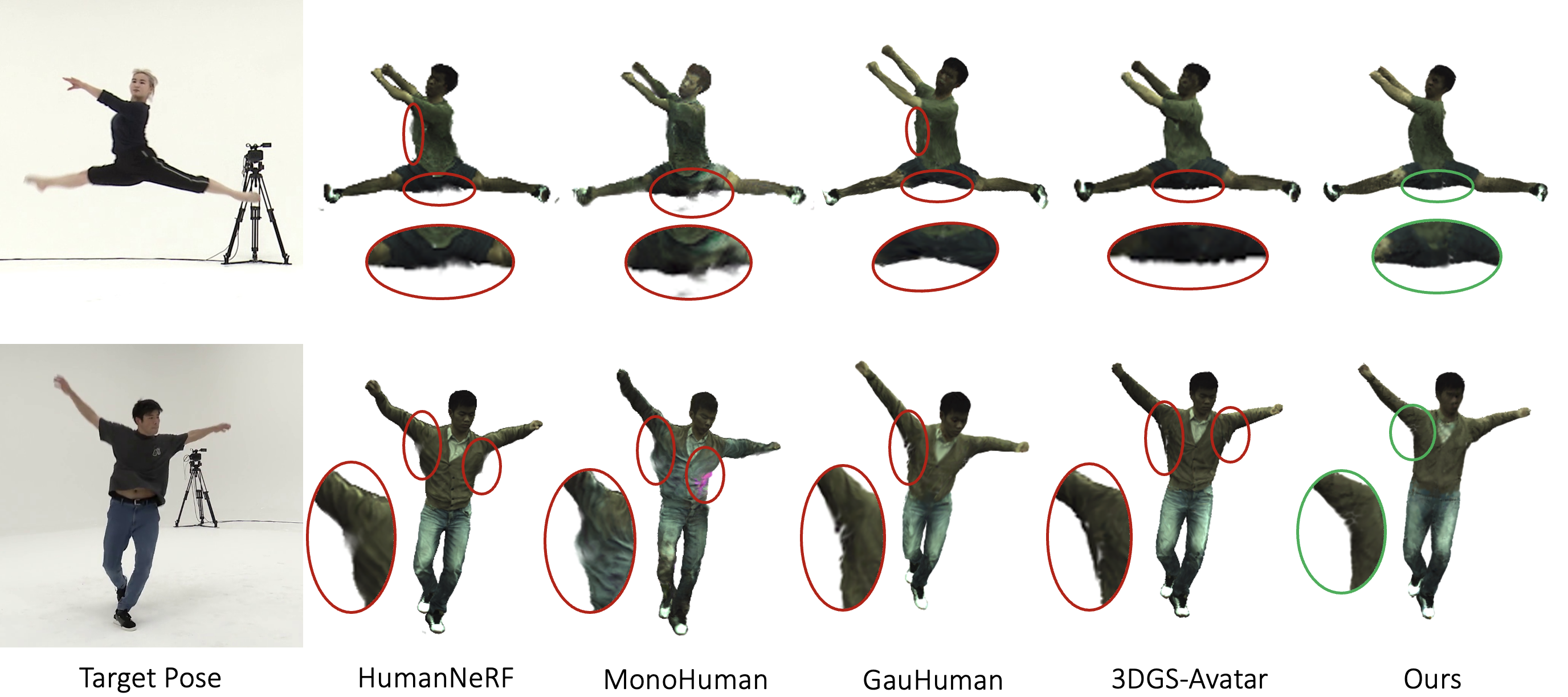}
 \vspace{-10pt}
  \caption{\textbf{Animation of human avatars under novel out-of-distribution poses.} Qualitative comparison with state-of-the-art approaches, namely HumanNeRF~\cite{weng2022humannerf}, MonoHuman~\cite{yu2023monohuman}, GauHuman~\cite{hu2023gauhuman}, and 3DGS-Avatar~\cite{qian20233dgs}. The training subjects are from the ZJU-MoCap dataset~\cite{peng2021neural} and the target poses (column 1) are from the AIST$++
  $ dataset~\cite{li2021learn,aist-dance-db}. Since the target poses are out of the training distribution (first 3 come from the advanced dance videos of AIST$++$), animation under these poses is challenging. Our method demonstrates significant robustness.
  }
  \label{fig:pose_transfer_zju}
  \vspace{-15pt}
\end{figure}

\section{Experiments}

\subsection{Datasets}\label{sec:data}

To evaluate our method, we use 2 datasets, namely ZJU-MoCap \cite{peng2021neural} and AIST$++$ \cite{li2021learn,aist-dance-db}. ZJU-MoCap is a widely used dataset that contains multi-view recordings of human subjects, performing motions such as kicking and arm swings. In our experiments, we assume a monocular setting and only use the camera 1 for training. Similarly with other works~\cite{mu2023actorsnerf,weng2022humannerf,yu2023monohuman,qian20233dgs}, we use 6 subjects (377, 386, 387, 392, 393, 394) and the corresponding camera parameters, SMPL~\cite{SMPL:2015} body poses, and segmentation masks provided by the dataset.

AIST$++$~\cite{li2021learn,aist-dance-db} is a dance motion dataset that includes a large variety of human body movements. In contrast to ZJU-MoCap that consists of mostly repetitive sequences with limited variation, AIST$++$ contains more diverse and challenging human body articulations. It contains 30 subjects and 10 dance genres. Each subject performs a specific number of \emph{basic} and \emph{advanced} dance choreographies. We randomly select 1 basic ($\sim$12 sec.) and 1 advanced ($\sim$30 sec.) video per subject. We assume a monocular setting by only using the camera 1 for training. To get more accurate and smooth tracking over time, we use 4DHumans~\cite{goel2023humans} to get the corresponding SMPL annotations, camera parameters, and segmentation masks for this dataset. We construct 2 test sets: (a) a ``Basic Test'' set that consists of 10 basic dance videos, and (b) an ``Advanced Test'' set of 10 advanced dance videos, specifically chosen to evaluate the robustness of our method to novel and more challenging (out-of-distribution) poses.

\subsection{Evaluation}\label{sec:evaluation}

\noindent
\textbf{Baselines.}  We compare our method against recent approaches for animatable human avatars, learned from monocular videos. We include both NeRF-based (HumanNeRF~\cite{weng2022humannerf} and MonoHuman~\cite{yu2023monohuman}) and 3DGS-based (GauHuman~\cite{hu2023gauhuman} and 3DGS-Avatar~\cite{qian20233dgs}) methods. HumanNeRF~\cite{weng2022humannerf} is a well-performing NeRF-based approach that achieves high-quality reconstruction of a human subject, given a monocular video. MonoHuman~\cite{yu2023monohuman} introduces a bi-directional deformation field to improve rendering of human subjects under novel poses. It requires keyframe selection of frontal and back views from a monocular video. GauHuman~\cite{hu2023gauhuman} and 3DGS-Avatar~\cite{qian20233dgs} are very recent approaches that apply 3DGS for humans, achieving fast optimization and real-time rendering. In our preliminary experiments, we observe that 3DGS-Avatar achieves higher-quality results than GauHuman, and thus we build our proposed multi-identity approach upon the single-identity 3DGS-Avatar, but omitting any per-frame latent codes to avoid any overfitting to the training frames (see \cref{sec:prelim_3dgs_humans}). We evaluate the performance of all the methods qualitatively and quantitatively when rendering a subject under novel poses and novel views.

\noindent
\textbf{Evaluation Metrics.}  We measure the visual quality of the generated videos, using standard reconstruction metrics, namely peak signal-to-noise ratio (PSNR), structural similarity index (SSIM)~\cite{wang2004image}, and learned perceptual image patch similarity (LPIPS)~\cite{zhang2018perceptual}. Following related works, we report $\text{LPIPS}^{*} = \text{LPIPS} \times 10^3$.

\begin{table}[tb]
  \caption{\textbf{Quantitative evaluation on ZJU-MoCap}. We compare our method with state-of-the-art approaches (HumanNeRF~\cite{weng2022humannerf}, MonoHuman~\cite{yu2023monohuman}, GauHuman~\cite{hu2023gauhuman}, 3DGS-Avatar~\cite{qian20233dgs}) on novel view synthesis, using the standard test set of ZJU-MoCap. We report PSNR, SSIM and $\text{LPIPS}^{*} = \text{LPIPS} \times 10^3$ on 4 subjects (377, 386, 392, 394).
  }
  \label{tab:zjumocap}
  \centering
  \scalebox{0.72}{
  \begin{tabular}{l|ccc|ccc|ccc|ccc}
    \toprule
    & \multicolumn{3}{c}{\textbf{377}} & \multicolumn{3}{c}{\textbf{386}}  & \multicolumn{3}{c}{\textbf{392}}  & \multicolumn{3}{c}{\textbf{394}} \\
    \textbf{Method} & PSNR$\uparrow$ & SSIM$\uparrow$ & LPIPS*$\downarrow$ & PSNR$\uparrow$ & SSIM$\uparrow$ & $\text{LPIPS}^{*}$$\downarrow$& PSNR$\uparrow$ & SSIM$\uparrow$ & $\text{LPIPS}^{*}$$\downarrow$& PSNR$\uparrow$ & SSIM$\uparrow$ & $\text{LPIPS}^{*}$$\downarrow$\\
    \midrule
    HumanNeRF & 30.41 & 0.9743 & 24.06 & 33.20 & 0.9752 & 28.99 & 31.04 & 0.9705 & 32.12 & 30.31 & 0.9642 & 32.89 \\
    MonoHuman & 29.12 & 0.9727 & 26.58 & 32.94 & 0.9695 & 36.04 & 29.50 & 0.9635 & 39.45 & 29.15 & 0.9595 & 38.08\\
    GauHuman & 29.48 &  0.9667 & 22.40 & 29.51 & 0.9648 & 28.80 & 27.86 & 0.9578 & 32.45 & 28.37 & 0.9533 & 34.16\\
    3DGS-Avatar & 30.64 & 0.9774 & 20.88 & 33.63 & 0.9773 & 25.77 & 31.66 & 0.9730 & 30.14 & 30.54 & 0.9661 & 31.21 \\
     Ours & \textbf{32.85} & \textbf{0.9847} & \textbf{19.67} & \textbf{34.98} & \textbf{0.9775} & \textbf{24.56} & \textbf{33.88} & \textbf{0.9805} & \textbf{25.73} & \textbf{32.28} & \textbf{0.9761} & \textbf{25.48} \\
  \bottomrule
  \end{tabular}
  }
  \vspace{-5pt}
\end{table}

\begin{figure}[tb]
  \centering
  \includegraphics[width=0.95\linewidth]{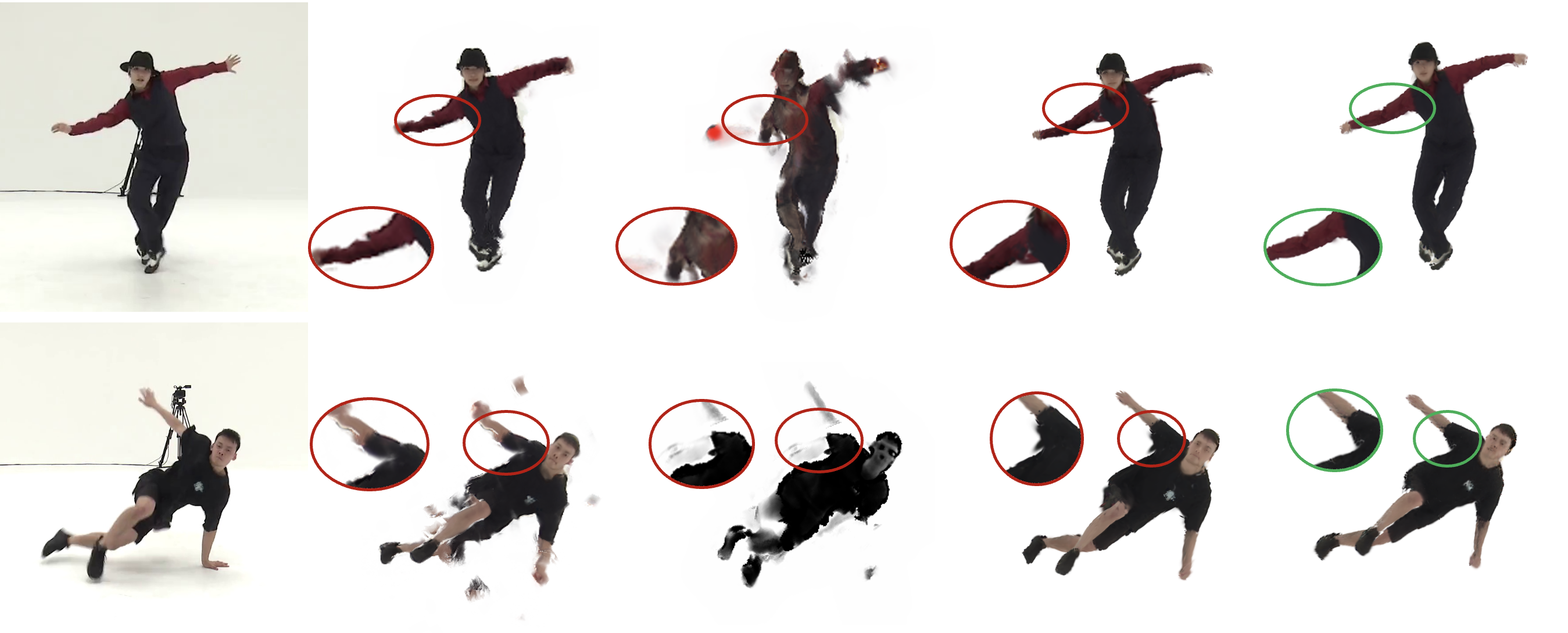}
\includegraphics[width=0.95\linewidth]{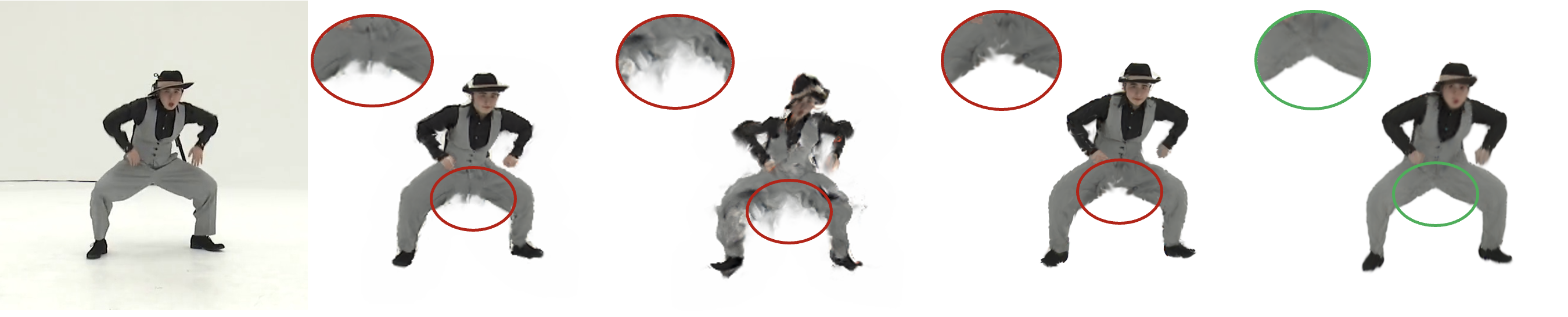}
\includegraphics[width=0.95\linewidth]{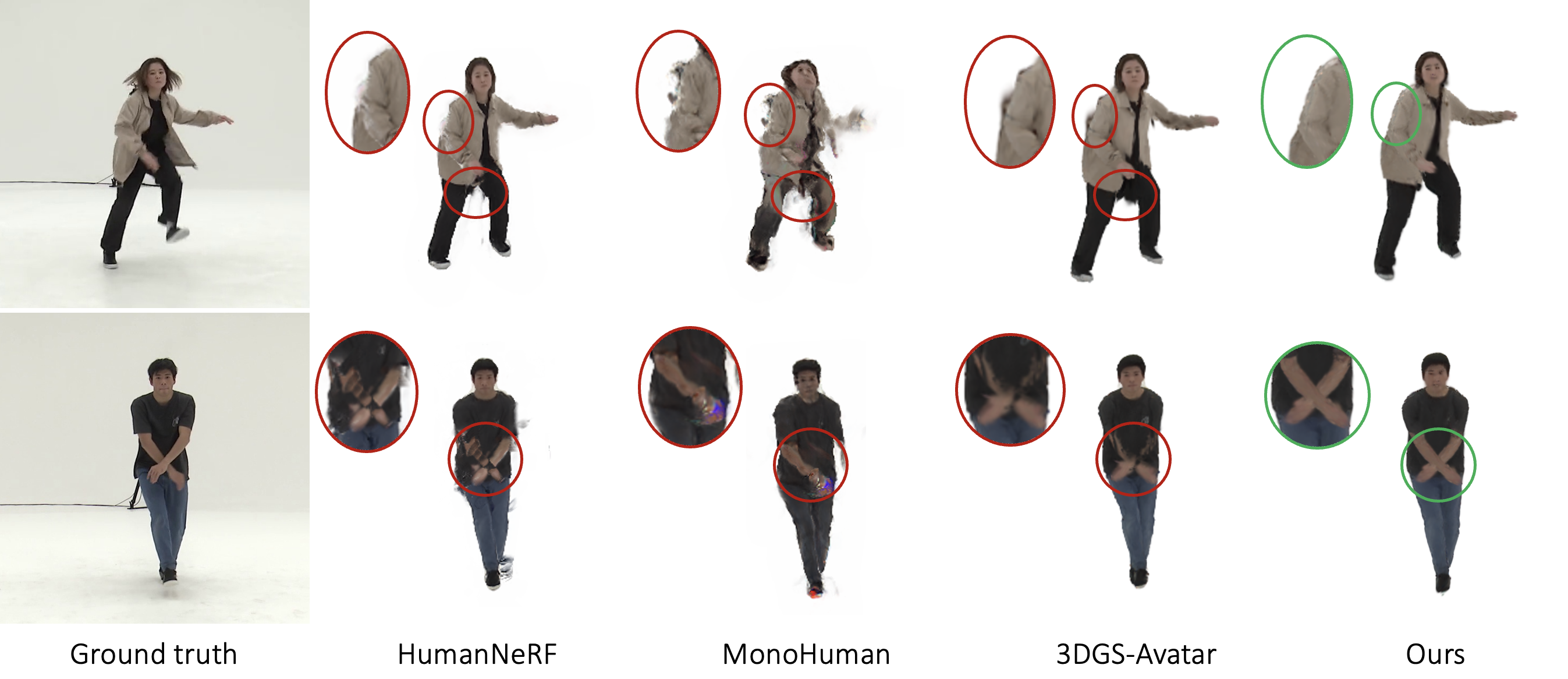}
  \caption{
  \textbf{Animation of human avatars under novel poses.} 
  Our method robustly animates all the identities under challenging unseen poses, from unseen camera views and advanced dance videos, outperforming the other methods, HumanNeRF~\cite{weng2022humannerf}, MonoHuman~\cite{yu2023monohuman}, and 3DGS-Avatar~\cite{qian20233dgs}. The subjects are from AIST++~\cite{li2021learn,aist-dance-db}. 
  }
  \label{fig:pose_aist}
  \vspace{-15pt}
\end{figure}


\subsubsection{Results on the ZJU-MoCap dataset.}

In our first evaluation setting, we train a model using the 6 subjects from ZJU-MoCap dataset~\cite{peng2021neural}. \cref{tab:zjumocap} shows the quantitative comparisons on novel view synthesis on the standard test set. We train all methods on the same training set (camera view 1) and test on the rest 22 camera views for each subject. For fair comparison, we do the same for GauHuman and report the corresponding values (in their original code they used the refined dataset of ZJU-MoCap). See suppl. for more comparisons.

We are mostly interested in the benefit of our multi-identity representation in animating the training subjects under novel poses. \Cref{fig:pose_transfer_zju} demonstrates the effectiveness of \MethodName in such a setting. More specifically, we test our model on target poses from the AIST$++$ dataset~\cite{li2021learn,aist-dance-db} (column 1 in \cref{fig:pose_transfer_zju}). These target poses are completely unseen and out of the training distribution, since in this case our model is trained on the 6 subjects from ZJU-MoCap. As mentioned in \cref{sec:data}, ZJU-MoCap includes more repetitive and common human motion compared to the dance sequences of AIST$++$. In addition, some of the target poses in \cref{fig:pose_transfer_zju} are from the advanced dance videos (first 3 rows). Thus, animating the subjects under those novel poses is more challenging. Artifacts under the arms and legs are visible in all other state-of-the-art methods, HumanNeRF~\cite{weng2022humannerf}, MonoHuman~\cite{yu2023monohuman}, GauHuman~\cite{hu2023gauhuman}, and 3DGS-Avatar~\cite{qian20233dgs}. Our proposed method demonstrates significant robustness, outperforming all the methods.

\begin{table}[tb]
  \caption{\textbf{Quantitative evaluation on AIST$++$.} We construct 2 test sets, a ``Basic Test'' and an ``Advanced Test'' set that consists of more advanced dance videos, specifically chosen to evaluate the robustness of our method to novel poses. We compare with state-of-the-art approaches (HumanNeRF~\cite{weng2022humannerf}, MonoHuman~\cite{yu2023monohuman}, 3DGS-Avatar~\cite{qian20233dgs}) and report PSNR, SSIM, and   $\text{LPIPS}^{*}=\text{LPIPS} \times 10^3$. Our method significantly outperforms the other methods.
  }
  \label{tab:aist}
  \centering
\scalebox{0.9}{
  \begin{tabular}{l|ccc|ccc}
    \toprule
    & \multicolumn{3}{c}{\textbf{Basic Test}} & \multicolumn{3}{c}{\textbf{Advanced Test}} \\
    \textbf{Method} & PSNR$\uparrow$ & SSIM$\uparrow$ & $\text{LPIPS}^{*}$$\downarrow$ & PSNR$\uparrow$ & SSIM$\uparrow$ & $\text{LPIPS}^{*}$$\downarrow$\\
    \midrule
    HumanNeRF~\cite{weng2022humannerf} & 24.58 & 0.9732  & 29.20 & 22.01 & 0.9654 & 39.01 \\
    MonoHuman~\cite{yu2023monohuman} & 21.02 & 0.8302  & 42.63 & 18.89 & 0.7540 & 85.50\\
    3DGS-Avatar~\cite{qian20233dgs} & 28.89 & 0.9846 & 18.20 & 25.51 & 0.9704 & 28.86 \\
    Ours & \textbf{29.82} & \textbf{0.9865} & \textbf{17.73} & \textbf{26.54} & \textbf{0.9741} & \textbf{26.02} \\
  \bottomrule
  \end{tabular}
  }
\end{table}

\subsubsection{Results on the AIST$++$ dataset.}

In our second evaluation setting, we train a model using all the 30 subjects of the AIST$++$ dataset~\cite{li2021learn,aist-dance-db}. Our training set includes both basic and advanced dance motion videos. We evaluate our model on both
our ``Basic Test'' and ``Advanced Test'' sets.
Our quantitative results are shown in \cref{tab:aist}. In both settings, \MethodName outperforms the existing approaches.

\Cref{fig:pose_aist} demonstrates the corresponding qualitative results, comparing with ground truth. The target poses (column 1) are unseen during training and come from unseen camera views. Poses in rows 2-4 are from challenging, more advanced dance combinations, and thus, are out of the training distribution. During training, the model has only seen more basic movements for those subjects. HumanNeRF~\cite{weng2022humannerf} and MonoHuman~\cite{yu2023monohuman} present significant artifacts in this case. Especially MonoHuman requires selection of frontal and back views, which might not be available in the monocular setting in the basic dance videos, leading to even wrong color (see first 2 rows in \cref{fig:pose_aist}). 3DGS-Avatar~\cite{qian20233dgs} is more robust but also produces artifacts in case of unseen limb articulations. Our method demonstrates robustness in animating all identities under arbitrary poses.

\begin{figure}[tb]
  \centering
    \includegraphics[width=\linewidth]{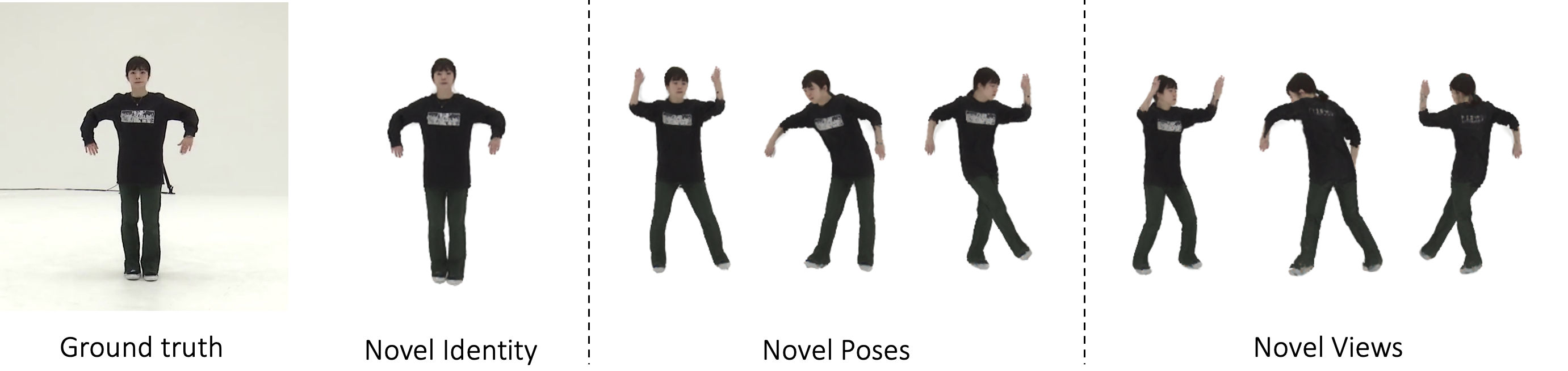}
  \caption{\textbf{Learning a novel identity.} Our method can be extended to learn a novel identity, and then animate it under novel poses and novel views.}
  \label{fig:novel_id}
  \vspace{-10pt}
\end{figure}

\Cref{fig:novel_id} shows the results of our method when learning a novel identity (see \cref{sec:pers_cont}), animating it under novel poses and rendering it under novel views. In this case, we learn an additional row in $\bm{U}_2$, given a short video of the new identity (10 sec.) with only basic movements. We optimize only the corresponding row of $\bm{U}_2$ and the color MLP $f_c$, keeping the rest of our representation frozen. 

\subsection{Ablation Study}

\begin{figure}[tb]
  \centering
  \begin{subfigure}{0.47\linewidth}
    \includegraphics[width=\linewidth]{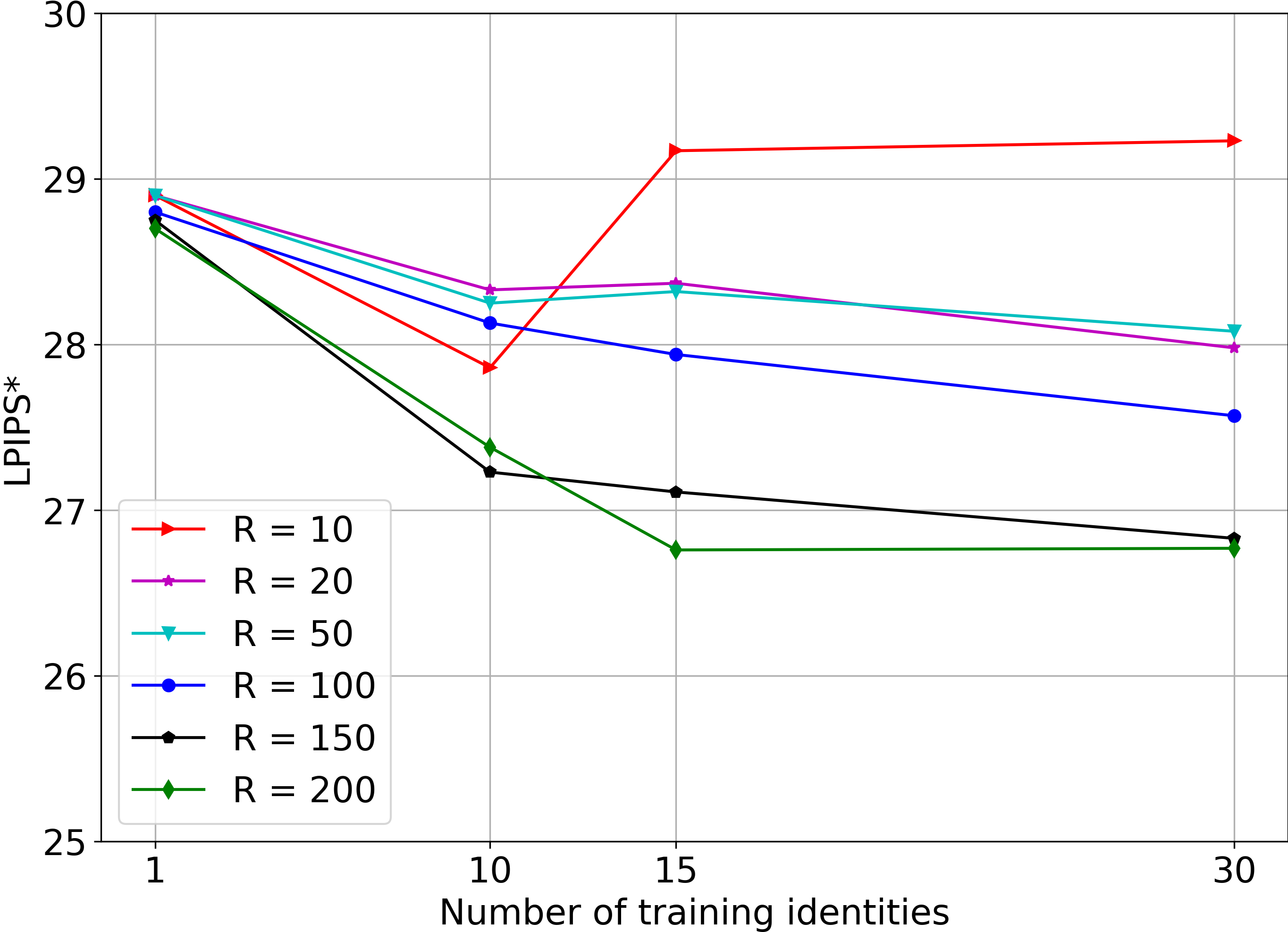}
    \caption{$\text{LPIPS}^{*}$ vs number of training identities}
    \label{fig:ablation_lpips_nids_R}
  \end{subfigure}
  \hfill
  \begin{subfigure}{0.47\linewidth}
    \includegraphics[width=\linewidth]{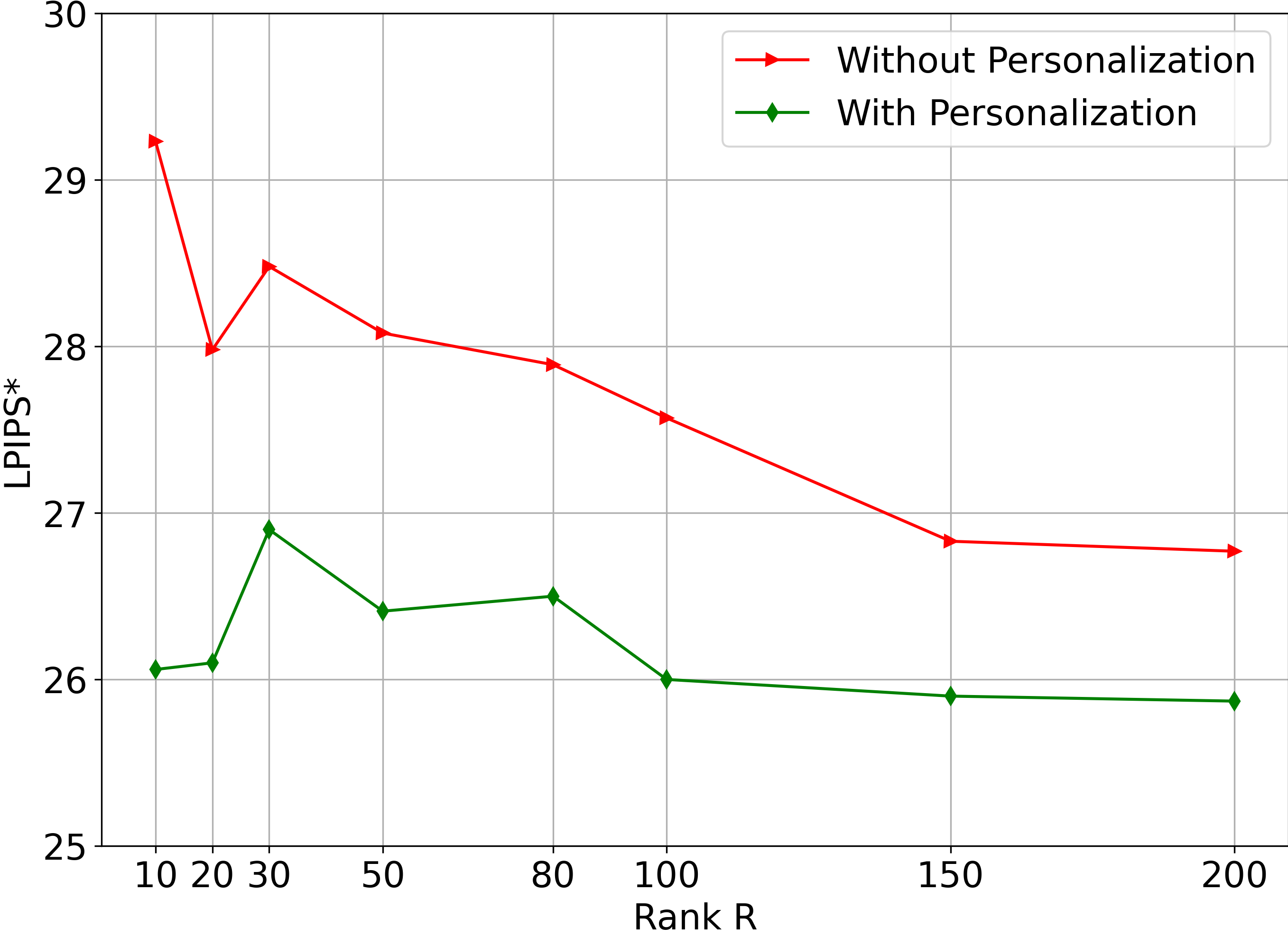}
    \caption{$\text{LPIPS}^{*}$ vs rank $R$}
    \label{fig:ablation_with_without}
  \end{subfigure}
  \vspace{-2pt}
  \caption{\textbf{Ablation study.}
  Visual quality ($\text{LPIPS}^{*}=\text{LPIPS} \times 10^3$) on the ``Advanced Test'' set for different values $R$ of our tensor decomposition. \textbf{(a)} $\text{LPIPS}^{*}$ (lower the better) vs the number of training identities (without any personalization). \textbf{(b)} $\text{LPIPS}^{*}$ vs rank $R$ for the model trained on 30 identities, with or without personalization. We observe that $R=10$ is not enough to capture a larger number of identities with diverse articulations, while $R=100$ is sufficient.}
  \label{fig:ablation}
  \vspace{-12pt}
\end{figure}

We conduct an ablation study on the rank $R$ of our tensor decomposition. \Cref{fig:ablation_lpips_nids_R} shows the average visual quality on the ``Advanced Test'' set for different number of training identities and for different values of $R$ (without any personalization). We observe that $R = 10$ is not enough to capture a larger number of identities. For larger values of $R$, we notice that increasing the number of identities leads to (a) an increase in robustness under novel poses (lower LPIPS), since our model leverages information from more subjects with diverse human body articulation, and (b) a smoother result, where our generic network might not capture high-frequency details. We address the described smoothing with our personalization procedure that, with only a few iterations, successfully captures such identity-specific details (see \cref{fig:ablation_with_without}). With personalization, we observe similar results for $R = 100$ and $R = 200$. We conclude that with just $R = 100$ for our learned tensor decomposition, we achieve robust animation under challenging novel poses. Further ablation studies and discussion about limitations and ethical considerations are included in the suppl.~document.


\section{Conclusion}

In conclusion, we introduce \MethodName, a novel method that learns a unified 3DGS representation for multiple identities. We propose to construct a high-order tensor that combines all the learnable parameters for multiple human avatars. We approximate this tensor with a tensor decomposition, significantly reducing the total number of learnable parameters. We quantitatively and qualitatively evaluate our method using a standard benchmark dataset and a more challenging dance motion dataset. By leveraging information from multiple identities, \MethodName enables robust animation of any training subject under novel challenging poses, out of the training distribution. In the future, we envision extending our approach to learn a generic model from thousands of identities.

\noindent
\textbf{Acknowledgements.} This work was partially supported by Amazon Prime Video. GC acknowledges travel support from the European Union’s Horizon 2020 research and innovation programme under grant agreement No 951847.

\bibliographystyle{splncs04}
\bibliography{main}

\begin{thebibliography}{100}
\providecommand{\url}[1]{\texttt{#1}}
\providecommand{\urlprefix}{URL }
\providecommand{\doi}[1]{https://doi.org/#1}

\bibitem{astrid2017cp}
Astrid, M., Lee, S.I.: Cp-decomposition with tensor power method for convolutional neural networks compression. In: 2017 IEEE International Conference on Big Data and Smart Computing (BigComp). pp. 115--118. IEEE (2017)

\bibitem{barron2021mipnerf}
Barron, J.T., Mildenhall, B., Tancik, M., Hedman, P., Martin-Brualla, R., Srinivasan, P.P.: Mip-nerf: A multiscale representation for anti-aliasing neural radiance fields (2021)

\bibitem{barron2022mipnerf360}
Barron, J.T., Mildenhall, B., Verbin, D., Srinivasan, P.P., Hedman, P.: Mip-nerf 360: Unbounded anti-aliased neural radiance fields. CVPR  (2022)

\bibitem{Bogo:ECCV:2016}
Bogo, F., Kanazawa, A., Lassner, C., Gehler, P., Romero, J., Black, M.J.: Keep it {SMPL}: Automatic estimation of {3D} human pose and shape from a single image. In: Computer Vision -- ECCV 2016. Lecture Notes in Computer Science, Springer International Publishing (Oct 2016)

\bibitem{8765346}
{Cao}, Z., {Hidalgo Martinez}, G., {Simon}, T., {Wei}, S., {Sheikh}, Y.A.: Openpose: Realtime multi-person 2d pose estimation using part affinity fields. IEEE Transactions on Pattern Analysis and Machine Intelligence  (2019)

\bibitem{cao2017realtime}
Cao, Z., Simon, T., Wei, S.E., Sheikh, Y.: Realtime multi-person 2d pose estimation using part affinity fields. In: CVPR (2017)

\bibitem{carranza2003free}
Carranza, J., Theobalt, C., Magnor, M.A., Seidel, H.P.: Free-viewpoint video of human actors. ACM transactions on graphics (TOG)  \textbf{22}(3),  569--577 (2003)

\bibitem{carroll1970analysis}
Carroll, J.D., Chang, J.J.: Analysis of individual differences in multidimensional scaling via an n-way generalization of “eckart-young” decomposition. Psychometrika  \textbf{35}(3),  283--319 (1970)

\bibitem{chan2019everybody}
Chan, C., Ginosar, S., Zhou, T., Efros, A.A.: Everybody dance now. In: Proceedings of the IEEE/CVF international conference on computer vision. pp. 5933--5942 (2019)

\bibitem{chen2021mvsnerf}
Chen, A., Xu, Z., Zhao, F., Zhang, X., Xiang, F., Yu, J., Su, H.: Mvsnerf: Fast generalizable radiance field reconstruction from multi-view stereo. arXiv preprint arXiv:2103.15595  (2021)

\bibitem{chen2021snarf}
Chen, X., Zheng, Y., Black, M.J., Hilliges, O., Geiger, A.: Snarf: Differentiable forward skinning for animating non-rigid neural implicit shapes. In: Proceedings of the IEEE/CVF International Conference on Computer Vision. pp. 11594--11604 (2021)

\bibitem{chrysos2021conditional}
Chrysos, G., Georgopoulos, M., Panagakis, Y.: Conditional generation using polynomial expansions. Advances in Neural Information Processing Systems  \textbf{34},  28390--28404 (2021)

\bibitem{de2008performance}
De~Aguiar, E., Stoll, C., Theobalt, C., Ahmed, N., Seidel, H.P., Thrun, S.: Performance capture from sparse multi-view video. In: ACM SIGGRAPH 2008 papers, pp. 1--10 (2008)

\bibitem{fridovich2022plenoxels}
Fridovich-Keil, S., Yu, A., Tancik, M., Chen, Q., Recht, B., Kanazawa, A.: Plenoxels: Radiance fields without neural networks. In: Proceedings of the IEEE/CVF Conference on Computer Vision and Pattern Recognition. pp. 5501--5510 (2022)

\bibitem{nerface}
Gafni, G., Thies, J., Zollhöfer, M., Nießner, M.: Dynamic neural radiance fields for monocular 4d facial avatar reconstruction (2020)

\bibitem{mpsnerf}
Gao, X., Yang, J., Kim, J., Peng, S., Liu, Z., Tong, X.: {MPS-NeRF: Generalizable 3D Human Rendering From Multiview Images}. IEEE Transactions on Pattern Analysis and Machine Intelligence  (2022)

\bibitem{georgopoulos2020multilinear}
Georgopoulos, M., Chrysos, G., Pantic, M., Panagakis, Y.: Multilinear latent conditioning for generating unseen attribute combinations. In: International Conference on Machine Learning. pp. 3442--3451. PMLR (2020)

\bibitem{10.1145/280814.280820}
Gleicher, M.: Retargetting motion to new characters. In: Proceedings of the 25th Annual Conference on Computer Graphics and Interactive Techniques. p. 33–42. SIGGRAPH '98, Association for Computing Machinery, New York, NY, USA (1998). \doi{10.1145/280814.280820}, \url{https://doi.org/10.1145/280814.280820}

\bibitem{goel2023humans}
Goel, S., Pavlakos, G., Rajasegaran, J., Kanazawa*, A., Malik*, J.: Humans in 4{D}: Reconstructing and tracking humans with transformers. In: International Conference on Computer Vision (ICCV) (2023)

\bibitem{goodfellow2014generative}
Goodfellow, I., Pouget-Abadie, J., Mirza, M., Xu, B., Warde-Farley, D., Ozair, S., Courville, A., Bengio, Y.: Generative adversarial nets. Advances in neural information processing systems  \textbf{27} (2014)

\bibitem{adnerf}
Guo, Y., Chen, K., Liang, S., Liu, Y.J., Bao, H., Zhang, J.: {AD-NeRF: Audio driven neural radiance fields for talking head synthesis}. In: IEEE International Conference on Computer Vision (ICCV). pp. 5784--5794 (2021)

\bibitem{harshman1970foundations}
Harshman, R.A., et~al.: Foundations of the parafac procedure: Models and conditions for an" explanatory" multimodal factor analysis  (1970)

\bibitem{hitchcock1927tensor_polyadic}
Hitchcock, F.L.: The expression of a tensor or a polyadic as a sum of products. J. of Math. and Phys.  \textbf{6}(1-4),  164--189 (1927). \doi{10.1002/sapm192761164}

\bibitem{hu2023gaussianavatar}
Hu, L., Zhang, H., Zhang, Y., Zhou, B., Liu, B., Zhang, S., Nie, L.: Gaussianavatar: Towards realistic human avatar modeling from a single video via animatable 3d gaussians. arXiv preprint arXiv:2312.02134  (2023)

\bibitem{hu2023gauhuman}
Hu, S., Liu, Z.: Gauhuman: Articulated gaussian splatting from monocular human videos. arXiv preprint arXiv:2312.02973  (2023)

\bibitem{pix2pix2017}
Isola, P., Zhu, J.Y., Zhou, T., Efros, A.A.: Image-to-image translation with conditional adversarial networks. CVPR  (2017)

\bibitem{jayasundara2023flexnerf}
Jayasundara, V., Agrawal, A., Heron, N., Shrivastava, A., Davis, L.S.: Flexnerf: Photorealistic free-viewpoint rendering of moving humans from sparse views. In: IEEE Conference on Computer Vision and Pattern Recognition (CVPR) (2023)

\bibitem{jiang2022neuman}
Jiang, W., Yi, K.M., Samei, G., Tuzel, O., Ranjan, A.: Neuman: Neural human radiance field from a single video. In: European Conference on Computer Vision. pp. 402--418. Springer (2022)

\bibitem{kanade1997constructing}
Kanade, T., Rander, P., Narayanan, P.: Constructing virtual worlds from real scenes. In: ACM Multimedia. vol.~1 (1997)

\bibitem{hmrKanazawa17}
Kanazawa, A., Black, M.J., Jacobs, D.W., Malik, J.: End-to-end recovery of human shape and pose. In: Computer Vision and Pattern Recognition (CVPR) (2018)

\bibitem{kania2022conerf}
Kania, K., Yi, K.M., Kowalski, M., Trzci{\'n}ski, T., Tagliasacchi, A.: {CoNeRF: Controllable Neural Radiance Fields}. In: IEEE Conference on Computer Vision and Pattern Recognition (CVPR) (2022)

\bibitem{kerbl20233d}
Kerbl, B., Kopanas, G., Leimk{\"u}hler, T., Drettakis, G.: 3d gaussian splatting for real-time radiance field rendering. ACM Transactions on Graphics  \textbf{42}(4) (2023)

\bibitem{khatri1968solutions}
Khatri, C., Rao, C.R.: Solutions to some functional equations and their applications to characterization of probability distributions. Sankhy{\=a}: The Indian Journal of Statistics, Series A pp. 167--180 (1968)

\bibitem{kim2023yoto}
Kim, J., Wee, D., Xu, D.: You only train once: Multi-identity free-viewpoint neural human rendering from monocular videos (2023), \url{https://arxiv.org/abs/2303.05835}

\bibitem{kingma2014adam}
Kingma, D.P., Ba, J.: Adam: A method for stochastic optimization. arXiv preprint arXiv:1412.6980  (2014)

\bibitem{kolda2009tensor}
Kolda, T.G., Bader, B.W.: Tensor decompositions and applications. SIAM review  \textbf{51}(3),  455--500 (2009)

\bibitem{kossaifi2017tensor}
Kossaifi, J., Khanna, A., Lipton, Z., Furlanello, T., Anandkumar, A.: Tensor contraction layers for parsimonious deep nets. In: Proceedings of the IEEE Conference on Computer Vision and Pattern Recognition Workshops. pp. 26--32 (2017)

\bibitem{kossaifi2020tensor}
Kossaifi, J., Lipton, Z.C., Kolbeinsson, A., Khanna, A., Furlanello, T., Anandkumar, A.: Tensor regression networks. Journal of Machine Learning Research  \textbf{21}(123),  1--21 (2020)

\bibitem{kossaifi2016tensorly}
Kossaifi, J., Panagakis, Y., Anandkumar, A., Pantic, M.: Tensorly: Tensor learning in python. arXiv preprint arXiv:1610.09555  (2016)

\bibitem{kwon2021neural}
Kwon, Y., Kim, D., Ceylan, D., Fuchs, H.: Neural human performer: Learning generalizable radiance fields for human performance rendering. Advances in Neural Information Processing Systems  \textbf{34},  24741--24752 (2021)

\bibitem{lebedev2014speeding}
Lebedev, V., Ganin, Y., Rakhuba, M., Oseledets, I., Lempitsky, V.: Speeding-up convolutional neural networks using fine-tuned cp-decomposition. arXiv preprint arXiv:1412.6553  (2014)

\bibitem{10.1145/311535.311539}
Lee, J., Shin, S.Y.: A hierarchical approach to interactive motion editing for human-like figures. In: Proceedings of the 26th Annual Conference on Computer Graphics and Interactive Techniques. p. 39–48. SIGGRAPH '99, ACM Press/Addison-Wesley Publishing Co., USA (1999). \doi{10.1145/311535.311539}, \url{https://doi.org/10.1145/311535.311539}

\bibitem{li2021learn}
Li, R., Yang, S., Ross, D.A., Kanazawa, A.: Learn to dance with aist++: Music conditioned 3d dance generation (2021)

\bibitem{li2022neural}
Li, T., Slavcheva, M., Zollhoefer, M., Green, S., Lassner, C., Kim, C., Schmidt, T., Lovegrove, S., Goesele, M., Newcombe, R., et~al.: Neural 3d video synthesis from multi-view video. In: Proceedings of the IEEE/CVF Conference on Computer Vision and Pattern Recognition. pp. 5521--5531 (2022)

\bibitem{li2021neural}
Li, Z., Niklaus, S., Snavely, N., Wang, O.: Neural scene flow fields for space-time view synthesis of dynamic scenes. In: Proceedings of the IEEE/CVF Conference on Computer Vision and Pattern Recognition. pp. 6498--6508 (2021)

\bibitem{lindell2022bacon}
Lindell, D.B., Van~Veen, D., Park, J.J., Wetzstein, G.: Bacon: Band-limited coordinate networks for multiscale scene representation. In: Proceedings of the IEEE/CVF Conference on Computer Vision and Pattern Recognition. pp. 16252--16262 (2022)

\bibitem{liu2021neuralactor}
Liu, L., Habermann, M., Rudnev, V., Sarkar, K., Gu, J., Theobalt, C.: Neural actor: Neural free-view synthesis of human actors with pose control. ACM Trans. Graph.(ACM SIGGRAPH Asia)  (2021)

\bibitem{SMPL:2015}
Loper, M., Mahmood, N., Romero, J., Pons-Moll, G., Black, M.J.: {SMPL}: A skinned multi-person linear model. ACM Transactions on Graphics, (Proc. SIGGRAPH Asia)  \textbf{34}(6),  248:1--248:16 (Oct 2015)

\bibitem{loper2023smpl}
Loper, M., Mahmood, N., Romero, J., Pons-Moll, G., Black, M.J.: Smpl: A skinned multi-person linear model. In: Seminal Graphics Papers: Pushing the Boundaries, Volume 2, pp. 851--866 (2023)

\bibitem{luiten2023dynamic}
Luiten, J., Kopanas, G., Leibe, B., Ramanan, D.: Dynamic 3d gaussians: Tracking by persistent dynamic view synthesis. In: 3DV (2024)

\bibitem{AMASS:ICCV:2019}
Mahmood, N., Ghorbani, N., Troje, N.F., Pons-Moll, G., Black, M.J.: {AMASS}: Archive of motion capture as surface shapes. In: International Conference on Computer Vision. pp. 5442--5451 (Oct 2019)

\bibitem{matusik2000image}
Matusik, W., Buehler, C., Raskar, R., Gortler, S.J., McMillan, L.: Image-based visual hulls. In: Proceedings of the 27th annual conference on Computer graphics and interactive techniques. pp. 369--374 (2000)

\bibitem{LEAP:CVPR:21}
Mihajlovic, M., Zhang, Y., Black, M.J., Tang, S.: {LEAP}: Learning articulated occupancy of people. In: Proceedings IEEE Conf. on Computer Vision and Pattern Recognition (CVPR) (Jun 2021)

\bibitem{mildenhall2020nerf}
Mildenhall, B., Srinivasan, P.P., Tancik, M., Barron, J.T., Ramamoorthi, R., Ng, R.: Nerf: Representing scenes as neural radiance fields for view synthesis. In: ECCV (2020)

\bibitem{mohlenkamp2013musings}
Mohlenkamp, M.J.: Musings on multilinear fitting. Linear Algebra and its Applications  \textbf{438}(2),  834--852 (2013)

\bibitem{mu2023actorsnerf}
Mu, J., Sang, S., Vasconcelos, N., Wang, X.: {ActorsNeRF:} animatable few-shot human rendering with generalizable nerfs pp. 18391--18401 (2023)

\bibitem{muller2022instant}
M{\"u}ller, T., Evans, A., Schied, C., Keller, A.: Instant neural graphics primitives with a multiresolution hash encoding. ACM Transactions on Graphics (ToG)  \textbf{41}(4),  1--15 (2022)

\bibitem{neverova2018dense}
Neverova, N., Guler, R.A., Kokkinos, I.: Dense pose transfer. In: Proceedings of the European conference on computer vision (ECCV). pp. 123--138 (2018)

\bibitem{2021narf}
Noguchi, A., Sun, X., Lin, S., Harada, T.: {Neural Articulated Radiance Field}. In: International Conference on Computer Vision (ICCV) (2021)

\bibitem{novikov2015tensorizing}
Novikov, A., Podoprikhin, D., Osokin, A., Vetrov, D.P.: Tensorizing neural networks. Advances in neural information processing systems  \textbf{28} (2015)

\bibitem{STAR:2020}
Osman, A.A.A., Bolkart, T., Black, M.J.: {STAR}: A sparse trained articulated human body regressor. In: European Conference on Computer Vision (ECCV). pp. 598--613 (2020), \url{https://star.is.tue.mpg.de}

\bibitem{Pan_2023_ICCV}
Pan, X., Yang, Z., Ma, J., Zhou, C., Yang, Y.: Transhuman: A transformer-based human representation for generalizable neural human rendering. In: Proceedings of the IEEE/CVF International Conference on Computer Vision (ICCV). pp. 3544--3555 (October 2023)

\bibitem{panagakis2021tensor}
Panagakis, Y., Kossaifi, J., Chrysos, G.G., Oldfield, J., Nicolaou, M.A., Anandkumar, A., Zafeiriou, S.: Tensor methods in computer vision and deep learning. Proceedings of the IEEE  \textbf{109}(5),  863--890 (2021)

\bibitem{nerfies}
Park, K., Sinha, U., Barron, J.T., Bouaziz, S., Goldman, D.B., Seitz, S.M., Martin-Brualla, R.: Nerfies: Deformable neural radiance fields. ICCV  (2021)

\bibitem{park2021hypernerf}
Park, K., Sinha, U., Hedman, P., Barron, J.T., Bouaziz, S., Goldman, D.B., Martin-Brualla, R., Seitz, S.M.: Hypernerf: A higher-dimensional representation for topologically varying neural radiance fields. arXiv preprint arXiv:2106.13228  (2021)

\bibitem{paszke2019pytorch}
Paszke, A., Gross, S., Massa, F., Lerer, A., Bradbury, J., Chanan, G., Killeen, T., Lin, Z., Gimelshein, N., Antiga, L., et~al.: Pytorch: An imperative style, high-performance deep learning library. Advances in neural information processing systems  \textbf{32} (2019)

\bibitem{SMPL-X:2019}
Pavlakos, G., Choutas, V., Ghorbani, N., Bolkart, T., Osman, A.A.A., Tzionas, D., Black, M.J.: Expressive body capture: 3d hands, face, and body from a single image. In: Proceedings IEEE Conf. on Computer Vision and Pattern Recognition (CVPR) (2019)

\bibitem{peng2023implicit}
Peng, S., Geng, C., Zhang, Y., Xu, Y., Wang, Q., Shuai, Q., Zhou, X., Bao, H.: {Implicit Neural Representations with Structured Latent Codes for Human Body Modeling}. IEEE Transactions on Pattern Analysis and Machine Intelligence  (2023)

\bibitem{peng2021neural}
Peng, S., Zhang, Y., Xu, Y., Wang, Q., Shuai, Q., Bao, H., Zhou, X.: Neural body: Implicit neural representations with structured latent codes for novel view synthesis of dynamic humans. In: CVPR (2021)

\bibitem{Prokudin_2021_WACV}
Prokudin, S., Black, M.J., Romero, J.: Smplpix: Neural avatars from 3d human models. In: Proceedings of the IEEE/CVF Winter Conference on Applications of Computer Vision (WACV). pp. 1810--1819 (January 2021)

\bibitem{pumarola2021d}
Pumarola, A., Corona, E., Pons-Moll, G., Moreno-Noguer, F.: D-nerf: Neural radiance fields for dynamic scenes. In: Proceedings of the IEEE/CVF Conference on Computer Vision and Pattern Recognition. pp. 10318--10327 (2021)

\bibitem{qian20233dgs}
Qian, Z., Wang, S., Mihajlovic, M., Geiger, A., Tang, S.: 3dgs-avatar: Animatable avatars via deformable 3d gaussian splatting. arXiv preprint arXiv:2312.09228  (2023)

\bibitem{raj2021pixel}
Raj, A., Zollhofer, M., Simon, T., Saragih, J., Saito, S., Hays, J., Lombardi, S.: Pixel-aligned volumetric avatars. In: Proceedings of the IEEE/CVF Conference on Computer Vision and Pattern Recognition. pp. 11733--11742 (2021)

\bibitem{rajasegaran2022tracking}
Rajasegaran, J., Pavlakos, G., Kanazawa, A., Malik, J.: Tracking people by predicting 3d appearance, location and pose. In: Proceedings of the IEEE/CVF Conference on Computer Vision and Pattern Recognition. pp. 2740--2749 (2022)

\bibitem{MANO:SIGGRAPHASIA:2017}
Romero, J., Tzionas, D., Black, M.J.: Embodied hands: Modeling and capturing hands and bodies together. ACM Transactions on Graphics, (Proc. SIGGRAPH Asia)  \textbf{36}(6) (Nov 2017)

\bibitem{schonberger2016structure}
Schonberger, J.L., Frahm, J.M.: Structure-from-motion revisited. In: Proceedings of the IEEE conference on computer vision and pattern recognition. pp. 4104--4113 (2016)

\bibitem{simon2017hand}
Simon, T., Joo, H., Matthews, I., Sheikh, Y.: Hand keypoint detection in single images using multiview bootstrapping. In: CVPR (2017)

\bibitem{su2021nerf}
Su, S.Y., Yu, F., Zollh{\"o}fer, M., Rhodin, H.: A-nerf: Articulated neural radiance fields for learning human shape, appearance, and pose. Advances in Neural Information Processing Systems  \textbf{34},  12278--12291 (2021)

\bibitem{trevithick2021grf}
Trevithick, A., Yang, B.: Grf: Learning a general radiance field for 3d representation and rendering. In: Proceedings of the IEEE/CVF International Conference on Computer Vision. pp. 15182--15192 (2021)

\bibitem{aist-dance-db}
Tsuchida, S., Fukayama, S., Hamasaki, M., Goto, M.: Aist dance video database: Multi-genre, multi-dancer, and multi-camera database for dance information processing. In: Proceedings of the 20th International Society for Music Information Retrieval Conference, {ISMIR} 2019. Delft, Netherlands (Nov 2019)

\bibitem{turk1991eigenfaces}
Turk, M., Pentland, A.: Eigenfaces for recognition. Journal of cognitive neuroscience  \textbf{3}(1),  71--86 (1991)

\bibitem{vasilescu2002multilinear}
Vasilescu, M.A.O., Terzopoulos, D.: Multilinear analysis of image ensembles: Tensorfaces. ECCV (1)  \textbf{2350},  447--460 (2002)

\bibitem{vlasic2008articulated}
Vlasic, D., Baran, I., Matusik, W., Popovi{\'c}, J.: Articulated mesh animation from multi-view silhouettes. In: Acm Siggraph 2008 papers, pp.~1--9 (2008)

\bibitem{wang2021ibrnet}
Wang, Q., Wang, Z., Genova, K., Srinivasan, P., Zhou, H., Barron, J.T., Martin-Brualla, R., Snavely, N., Funkhouser, T.: Ibrnet: Learning multi-view image-based rendering. In: CVPR (2021)

\bibitem{wang2022arah}
Wang, S., Schwarz, K., Geiger, A., Tang, S.: Arah: Animatable volume rendering of articulated human sdfs. In: European conference on computer vision. pp. 1--19. Springer (2022)

\bibitem{wang2022neural}
Wang, T., Sarafianos, N., Yang, M.H., Tung, T.: Neural rendering of humans in novel view and pose from monocular video. arXiv preprint arXiv:2204.01218  (2022)

\bibitem{wang2018video}
Wang, T.C., Liu, M.Y., Zhu, J.Y., Liu, G., Tao, A., Kautz, J., Catanzaro, B.: Video-to-video synthesis. arXiv preprint arXiv:1808.06601  (2018)

\bibitem{wang2018high}
Wang, T.C., Liu, M.Y., Zhu, J.Y., Tao, A., Kautz, J., Catanzaro, B.: High-resolution image synthesis and semantic manipulation with conditional gans. In: Proceedings of the IEEE conference on computer vision and pattern recognition. pp. 8798--8807 (2018)

\bibitem{wang2004image}
Wang, Z., Bovik, A.C., Sheikh, H.R., Simoncelli, E.P.: Image quality assessment: from error visibility to structural similarity. IEEE transactions on image processing  \textbf{13}(4),  600--612 (2004)

\bibitem{wei2016cpm}
Wei, S.E., Ramakrishna, V., Kanade, T., Sheikh, Y.: Convolutional pose machines. In: CVPR (2016)

\bibitem{weng2020vid2actor}
Weng, C.Y., Curless, B., Kemelmacher-Shlizerman, I.: Vid2actor: Free-viewpoint animatable person synthesis from video in the wild. arXiv preprint arXiv:2012.12884  (2020)

\bibitem{weng2022humannerf}
Weng, C.Y., Curless, B., Srinivasan, P.P., Barron, J.T., Kemelmacher-Shlizerman, I.: Humannerf: Free-viewpoint rendering of moving people from monocular video. In: Proceedings of the IEEE/CVF conference on computer vision and pattern Recognition. pp. 16210--16220 (2022)

\bibitem{xu2021h}
Xu, H., Alldieck, T., Sminchisescu, C.: {H-nerf: Neural radiance fields for rendering and temporal reconstruction of humans in motion}. Advances in Neural Information Processing Systems  \textbf{34},  14955--14966 (2021)

\bibitem{yu2021pixelnerf}
Yu, A., Ye, V., Tancik, M., Kanazawa, A.: {pixelNeRF}: Neural radiance fields from one or few images. In: CVPR (2021)

\bibitem{yu2023monohuman}
Yu, Z., Cheng, W., Liu, X., Wu, W., Lin, K.Y.: Monohuman: Animatable human neural field from monocular video. In: Proceedings of the IEEE/CVF Conference on Computer Vision and Pattern Recognition. pp. 16943--16953 (2023)

\bibitem{zhang2018perceptual}
Zhang, R., Isola, P., Efros, A.A., Shechtman, E., Wang, O.: The unreasonable effectiveness of deep features as a perceptual metric. In: CVPR (2018)

\bibitem{zheng2023gpsgaussian}
Zheng, S., Zhou, B., Shao, R., Liu, B., Zhang, S., Nie, L., Liu, Y.: Gps-gaussian: Generalizable pixel-wise 3d gaussian splatting for real-time human novel view synthesis. arXiv  (2023)

\bibitem{Zielonka2023Drivable3D}
Zielonka, W., Bagautdinov, T., Saito, S., Zollhöfer, M., Thies, J., Romero, J.: Drivable 3d gaussian avatars  (2023)

\bibitem{zielonka2023insta}
Zielonka, W., Bolkart, T., Thies, J.: Instant volumetric head avatars. In: Proceedings of the IEEE/CVF Conference on Computer Vision and Pattern Recognition. pp. 4574--4584 (2023)

\bibitem{zwicker2001ewa}
Zwicker, M., Pfister, H., Van~Baar, J., Gross, M.: Ewa volume splatting. In: Proceedings Visualization, 2001. VIS'01. pp. 29--538. IEEE (2001)

\end{thebibliography}

\clearpage
\setcounter{section}{0}
\renewcommand\thesection{\Alph{section}}

\section*{Appendix}

\noindent
The appendix is organized as follows: 
\begin{itemize}
    \item Additional Ablation Study in~\cref{sec:1}.
    \item Additional Results in~\cref{sec:2}.
    \item Implementation Details in~\cref{sec:3}.
    \item Limitations in~\cref{sec:5}.
    \item Ethical Considerations in~\cref{sec:6}.
\end{itemize}
We strongly encourage the readers to watch our supplementary video on our project page: \url{https://aggelinacha.github.io/MIGS/}.

\begin{figure}[b]
  \centering
  \begin{subfigure}{0.45\linewidth}
    \includegraphics[width=\linewidth]{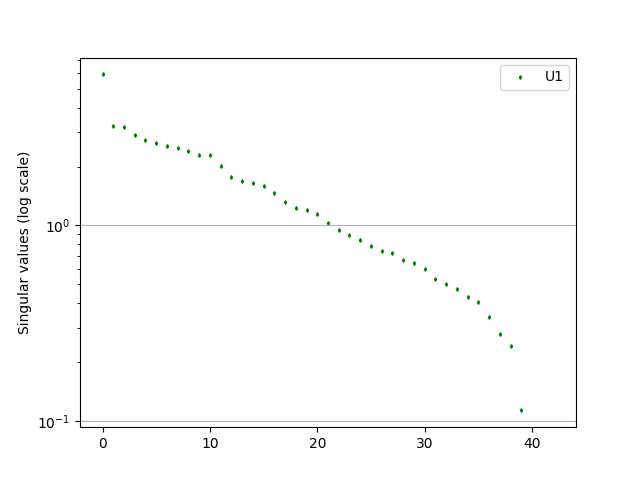}
    \caption{Singular values of $\bm{U}_1$}
    \label{fig:svd_U1}
  \end{subfigure}
  \hfill
  \begin{subfigure}{0.45\linewidth}
    \includegraphics[width=\linewidth]{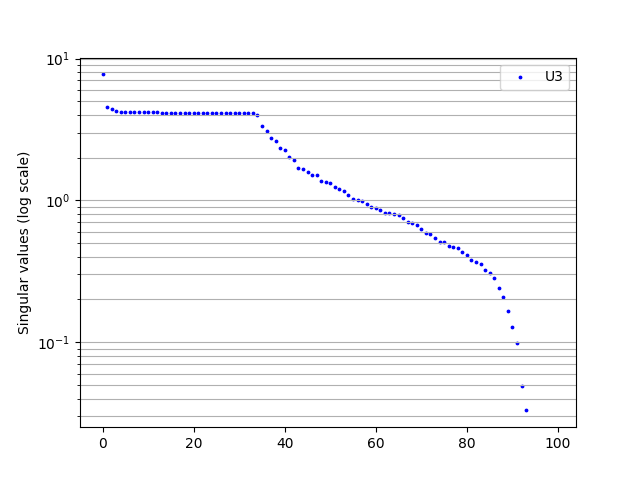}
    \caption{Singular values of $\bm{U}_3$}
    \label{fig:svd_U3}
  \end{subfigure}
  \caption{\textbf{Singular values.} The singular values of our learned $\bm{U}_1 \in \realnum ^{M \times R}$ and $\bm{U}_3 \in \realnum ^{N_g \times R}$, where $M = 43$ is the number of parameters per Gaussian, $N_g = 5 \times 10^4 $ is the number of 3D Gaussians, and $R = 100$ is the rank of our CP tensor decomposition.}
  \label{fig:svd}
  \vspace{-10pt}
\end{figure}

\section{Additional Ablation Study}\label{sec:1}

\noindent
\textbf{Learned Matrices.}
In \cref{fig:svd}, we plot the singular values of our learned matrices $\bm{U}_1 \in \realnum ^{M \times R}$ and $\bm{U}_3 \in \realnum ^{N_g \times R}$, where $M = 43$ is the number of parameters per Gaussian, $N_g = 5 \times 10^4 $ is the number of the 3D Gaussians, and $R = 100$ is the rank of our CP tensor decomposition (see Sec.~4.2 of the main paper). We use SVD from PyTorch~\cite{paszke2019pytorch}~\footnote{https://pytorch.org/docs/stable/generated/torch.svd.html} and plot them in logarithmic scale.
These matrices correspond to the model trained on the 30 subjects from the AIST++ dataset~\cite{li2021learn,aist-dance-db}.

\begin{figure}[tb]
  \centering
  \begin{subfigure}{0.45\linewidth}
    \includegraphics[width=\linewidth]{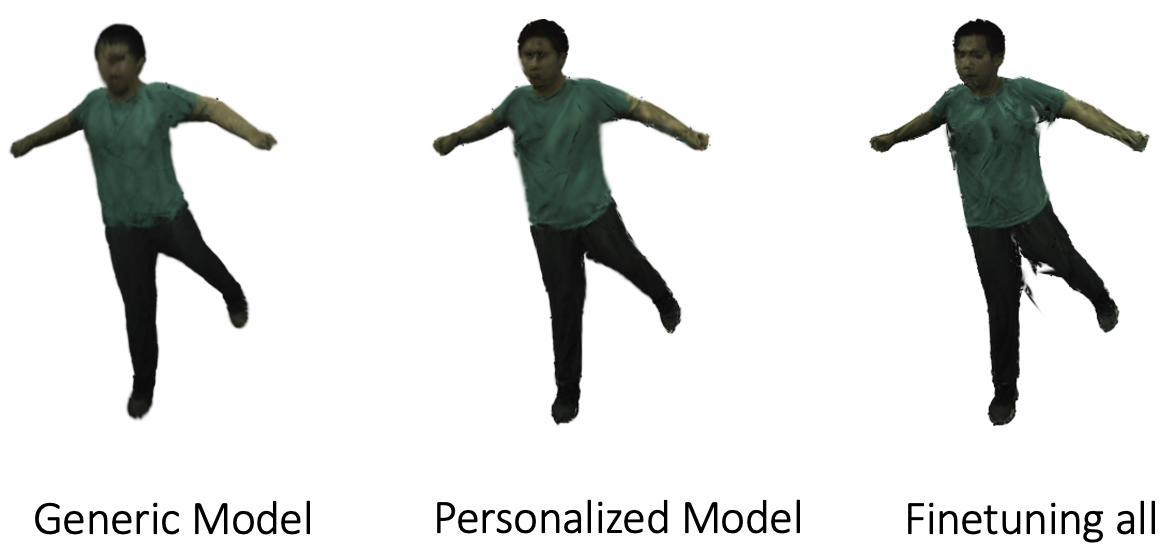}
    \caption{}
    \label{fig:ablation_personal}
  \end{subfigure}
  \hfill
  \begin{subfigure}{0.47\linewidth}
    \includegraphics[width=\linewidth]{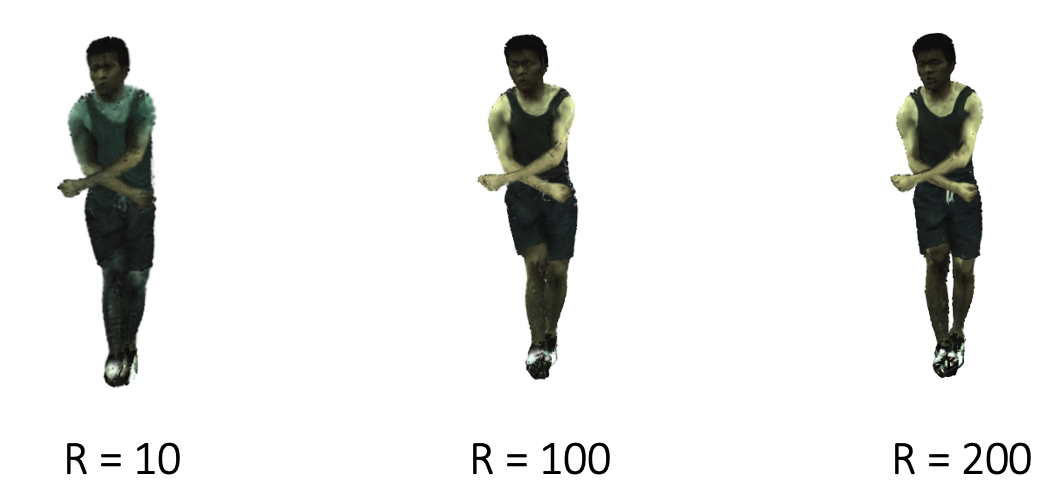}
    \caption{}
    \label{fig:ablation_R}
  \end{subfigure}
  \caption{\textbf{Ablation study.} \textbf{(a)} Ablation study on our proposed personalization procedure. In order to capture individual details, we fine-tune our color MLP of our generic model (left), leading to the personalized model for a particular subject (middle). We do not fine-tune all parameters (see artifacts on the right). \textbf{(b)} Ablation study on the rank $R$ of our tensor decomposition. $R=10$ leads to a mixture of identities (notice different shirt colors), while $R=100$ is sufficient to capture the training identities.}
  \label{fig:ablation_suppl}
\end{figure}

\begin{figure}[t]
  \centering
  \begin{subfigure}{0.45\linewidth}
    \includegraphics[width=\linewidth]{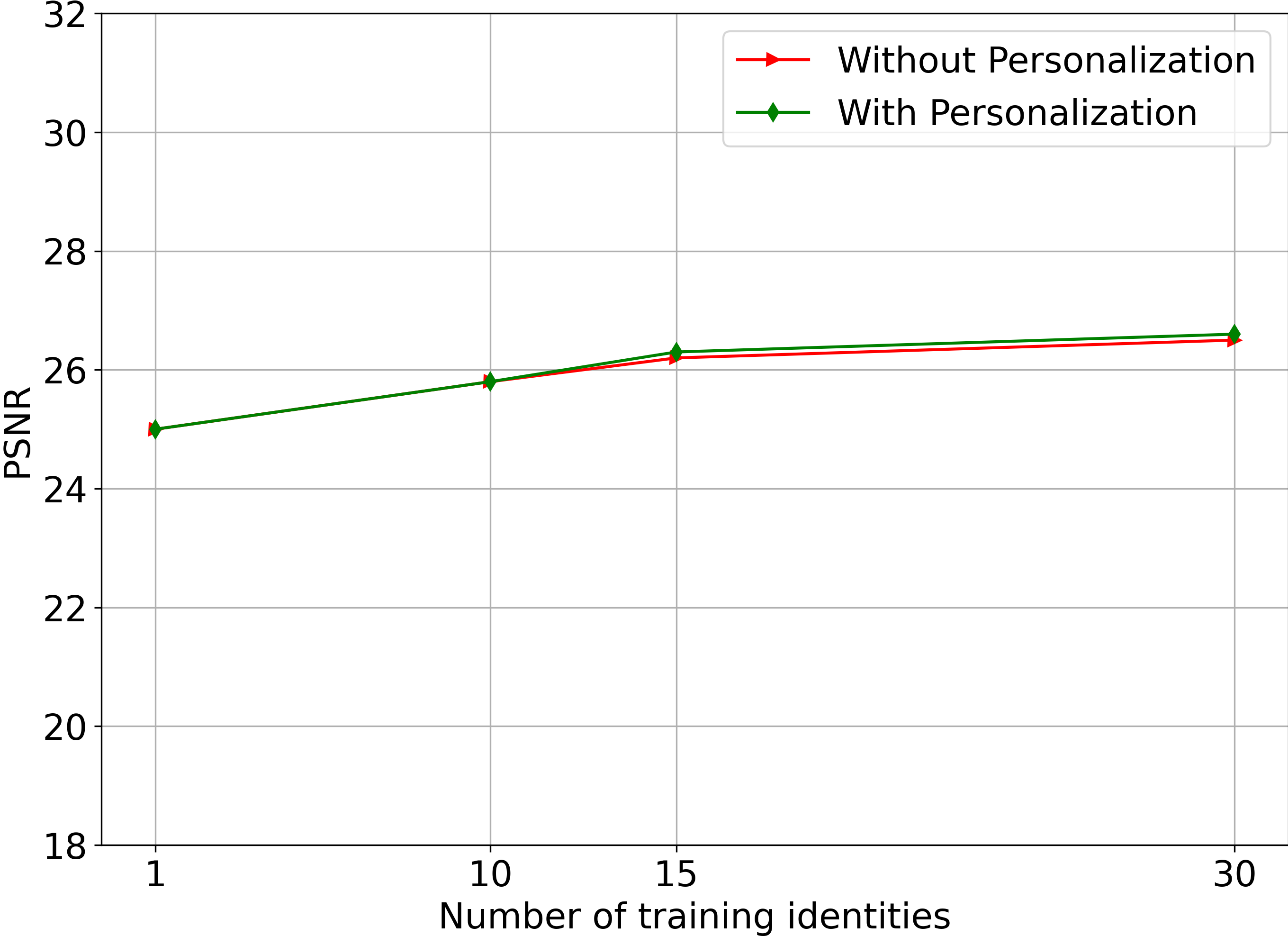}
    \caption{PSNR vs number of training identities}
    \label{fig:psnr_withwithout_Nids}
  \end{subfigure}
  \hfill
  \begin{subfigure}{0.45\linewidth}
    \includegraphics[width=\linewidth]{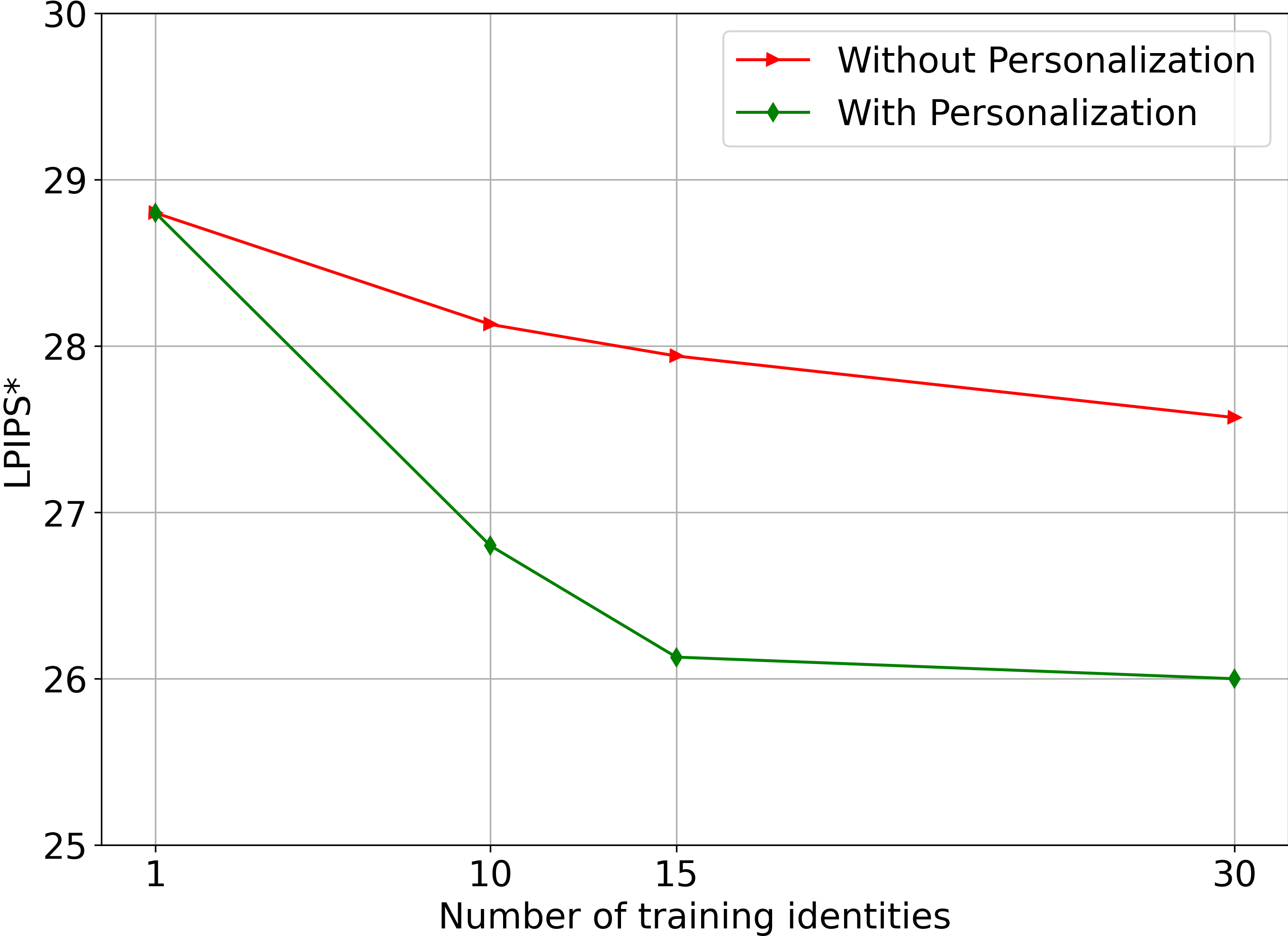}
    \caption{$\text{LPIPS}^{*}$ vs number of training identities}
    \label{fig:lpips_withwithout_Nids}
  \end{subfigure}
  \caption{\textbf{Ablation study.} Visual quality (PSNR and $\text{LPIPS}^{*}$) on the ``Advanced Test'' set, with or without personalization, for different number of training identities and $R=100$. }
  \label{fig:withwithout_a_b}
  \vspace{-10pt}
\end{figure}

\noindent
\textbf{Personalization.} As mentioned in Sec.~4.4 of the main paper, we can personalize our generic model for a particular subject, in order to capture individual details (\eg face or cloth details). We show an example output of our generic model trained on 30 identities and the corresponding personalized result in \cref{fig:ablation_personal}. We notice that this personalization procedure is needed only for models trained with identities more than 20. We only fine-tune the color MLP for $5 \times 10^3$ iterations, keeping the rest of the parameters frozen, using a short video of the target subject. We do not fine-tune all parameters, since in this case our network would forget the large variety of human body deformations learned from multiple subjects, leading to artifacts in novel poses (see \cref{fig:ablation_personal} right).

\noindent
\textbf{Ablation Study on the Rank $R$.} In \cref{fig:ablation_R}, we show a qualitative comparison for different values $R$ of our tensor decomposition (see also Sec.~5.3). We observe that $R=10$ is not enough to capture a larger number of identities. For example, it can lead to a mixture of colors (notice the shirt colors for $R=10$). On the other hand, $R = 100$ seems sufficient to capture all the training identities. 

\noindent
\textbf{Ablation Study on the Number of Identities.}
\Cref{fig:withwithout_a_b} shows the visual quality for different number of training identities, with and without personalization. As also mentioned in the main paper (see also Fig.~6), increasing the number of identities leads to an increase in robustness under novel poses. Further personalization captures identity-specific details, enhancing the output visual quality.

\begin{figure}[tb]
  \centering
    \includegraphics[width=\linewidth]{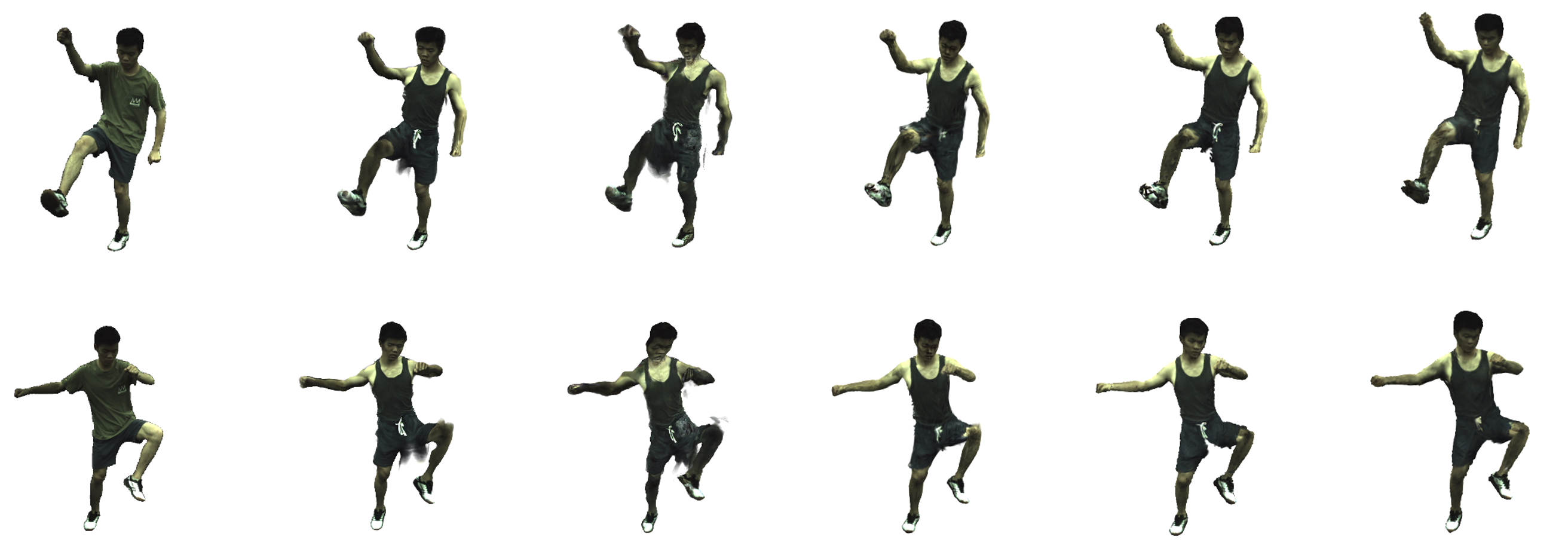}
     \includegraphics[width=\linewidth]{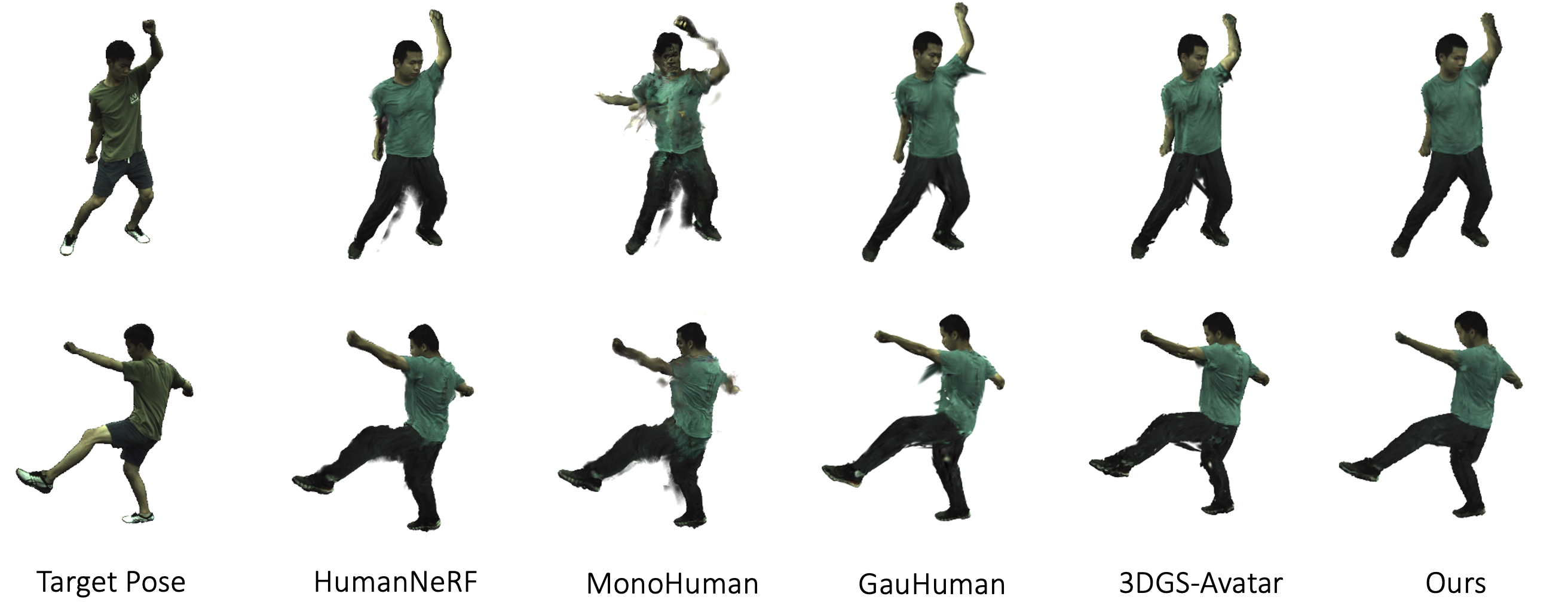}
  \caption{\textbf{Animation of human avatars under novel poses.} Qualitative comparison with state-of-the-art approaches, namely HumanNeRF~\cite{weng2022humannerf}, MonoHuman~\cite{yu2023monohuman}, GauHuman~\cite{hu2023gauhuman}, and 3DGS-Avatar~\cite{qian20233dgs}. The training subjects and the target poses are from the ZJU-MoCap dataset~\cite{peng2021neural}. Our method demonstrates significant robustness.
  }
  \label{fig:pose_transfer_zju}
\end{figure}

\begin{figure}[tb]
  \centering
    \includegraphics[width=\linewidth]{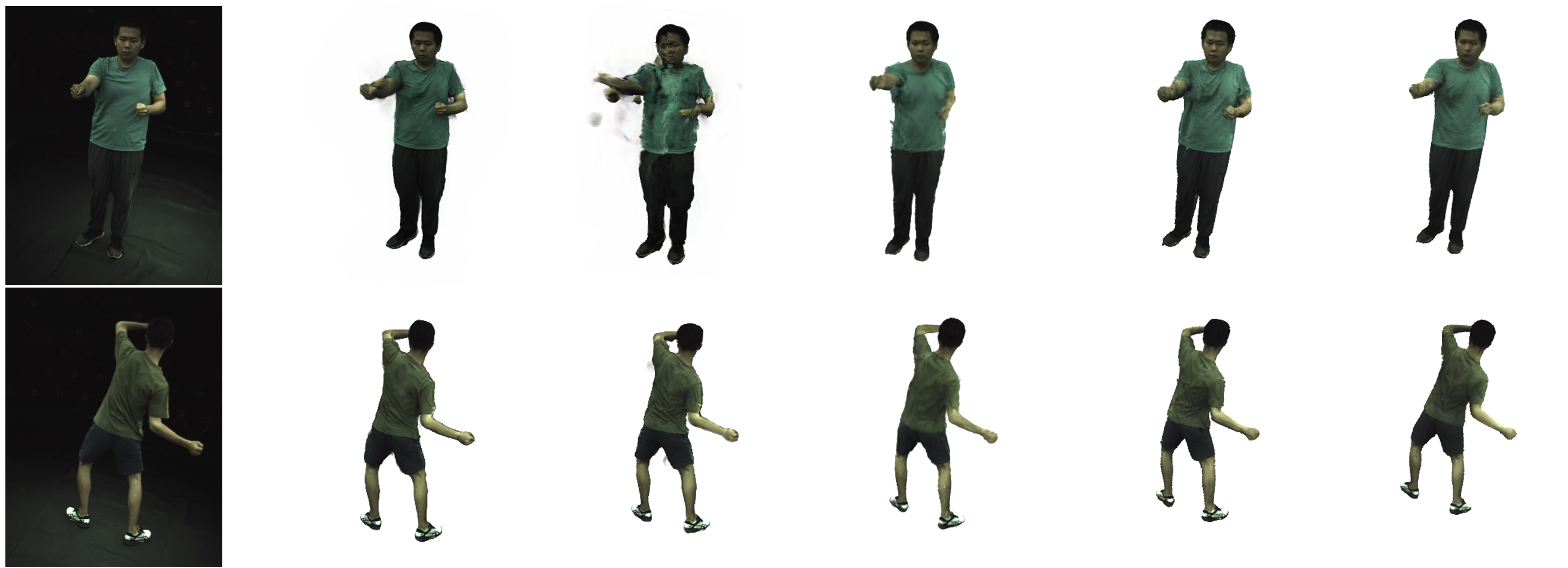}
     \includegraphics[width=\linewidth]{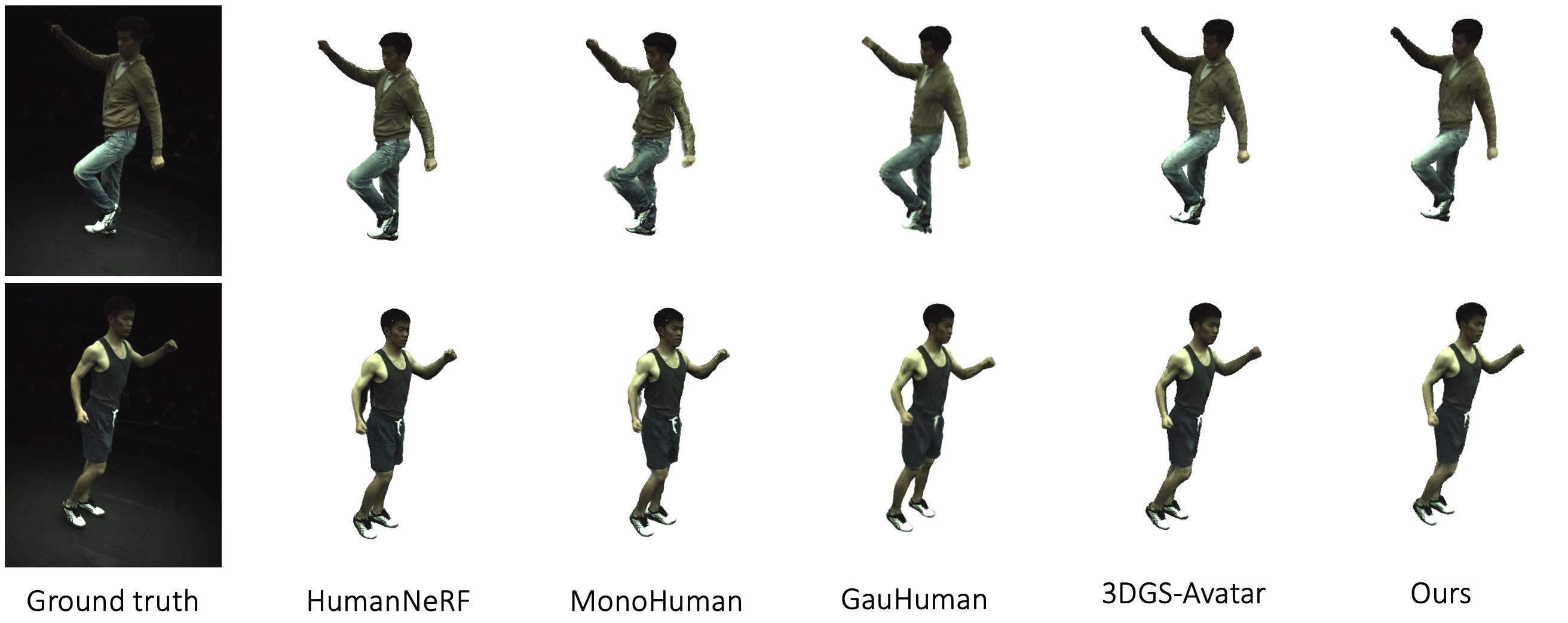}
  \caption{\textbf{Novel view synthesis on ZJU-MoCap}. Qualitative comparison with state-of-the-art approaches, namely HumanNeRF~\cite{weng2022humannerf}, MonoHuman~\cite{yu2023monohuman}, GauHuman~\cite{hu2023gauhuman}, and 3DGS-Avatar~\cite{qian20233dgs} on novel view synthesis on the test set of the ZJU-MoCap dataset~\cite{peng2021neural}.}
  \label{fig:novelview_zju}
\end{figure}

\begin{table}[tb]
  \caption{\textbf{Quantitative evaluation on ZJU-MoCap}. We compare our method with state-of-the-art approaches (HumanNeRF~\cite{weng2022humannerf}, MonoHuman~\cite{yu2023monohuman}, GauHuman~\cite{hu2023gauhuman}, 3DGS-Avatar~\cite{qian20233dgs}) on novel view synthesis, using the standard test set of ZJU-MoCap. We report PSNR, SSIM, and $\text{LPIPS}^{*} = \text{LPIPS} \times 10^3$ on 2 subjects (387, 393). See Table 1 in the main paper for the other 4 subjects (377, 386, 392, 394).
  }
  \label{tab:zjumocap_suppl}
  \centering
  \scalebox{0.72}{
  \begin{tabular}{l|ccc|ccc}
    \toprule
    & \multicolumn{3}{c}{\textbf{387}} & \multicolumn{3}{c}{\textbf{393}}  \\
    \textbf{Method} & PSNR$\uparrow$ & SSIM$\uparrow$ & $\text{LPIPS}^{*}$$\downarrow$ & PSNR$\uparrow$ & SSIM$\uparrow$ & $\text{LPIPS}^{*}$$\downarrow$\\
    \midrule
    HumanNeRF & 28.18 & 0.9632 & 35.58 & 28.31 & 0.9603 & 36.72 \\
    MonoHuman & 27.93 & 0.9601 & 41.76 & 27.64 & 0.9566 & 43.17 \\
    GauHuman & 27.95 &  0.9608 & 40.70 & 27.88 & 0.9578 & 43.01 \\
    3DGS-Avatar & 28.33 & 0.9642 & \textbf{34.24} & 28.88 & 0.9635 & 35.26 \\
    Ours & \textbf{30.70} & \textbf{0.9643} & 35.33 & \textbf{31.57} & \textbf{0.9640} & \textbf{30.44} \\
    
  \bottomrule
  \end{tabular}
  }
\end{table}

\begin{figure}[t]
  \centering
  \includegraphics[width=0.7\linewidth]{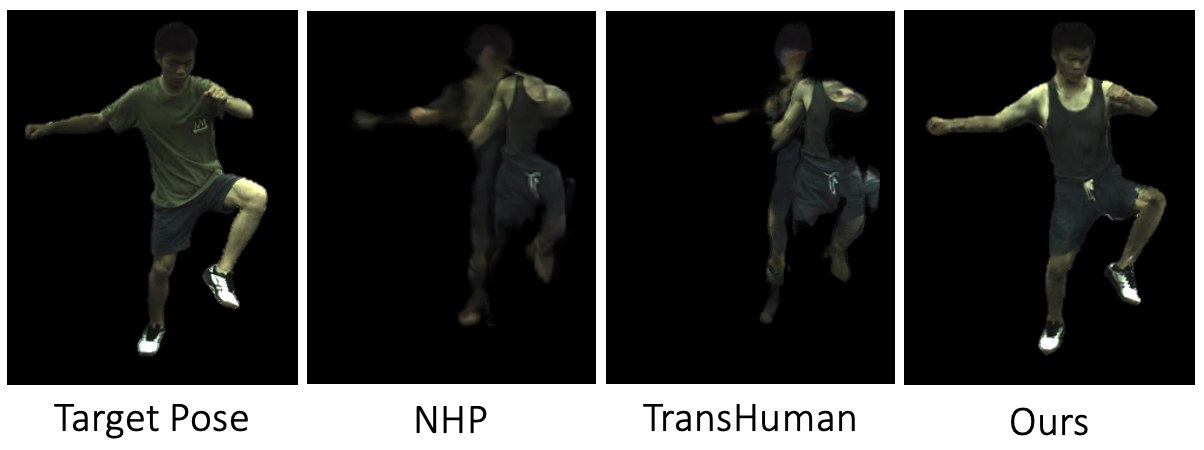}
   \caption{Comparison with multi-view approaches: Neural Human Performer~\cite{kwon2021neural} and TransHuman~\cite{Pan_2023_ICCV}.}
   \label{fig:multiview}
     \vspace{-10pt}
\end{figure}

\section{Additional Results}\label{sec:2}

\Cref{fig:pose_transfer_zju} demonstrates additional qualitative results for our model trained on subjects from the ZJU-MoCap dataset~\cite{peng2021neural}. Animating them under novel poses leads to artifacts under the arms and legs in all other methods, namely HumanNeRF~\cite{weng2022humannerf}, MonoHuman~\cite{yu2023monohuman}, GauHuman~\cite{hu2023gauhuman}, and 3DGS-Avatar~\cite{qian20233dgs}. Our proposed method demonstrates significant robustness.

\Cref{fig:pose_aist_suppl} demonstrates additional qualitative results when the target poses are from the AIST++ dataset~\cite{li2021learn,aist-dance-db}. In this case, the poses are more challenging, completely unseen during training (out of the training distribution). Again, \MethodName outperforms the other methods, robustly animating the identities under novel poses.

\Cref{fig:novelview_zju} shows qualitative comparisons for novel view synthesis on the test set of the ZJU-MoCap dataset~\cite{peng2021neural}. Corresponding quantitative results are shown in Table 1 of the main paper and \cref{tab:zjumocap_suppl}. Our method demonstrates comparable performance with 3DGS-Avatar on novel view synthesis, while trained on multiple identities simultaneously.

\Cref{fig:multiview} shows qualitative comparisons with approaches that use multiple views as input. These works address the problem in a different way. They extract features from nearby views and infer a novel view in a feed-forward manner. Most of them cannot animate humans under novel poses, \eg see results of NHP~\cite{li2021neural} and TransHuman~\cite{Pan_2023_ICCV} (we use black background to be consistent with their pretrained models - compare with row 2 in \cref{fig:pose_transfer_zju}). ActorsNeRF~\cite{mu2023actorsnerf} does not have code available. GPS-Gaussian~\cite{zheng2023gpsgaussian} does not use body pose (SMPL parameters) as input at all, and thus cannot render novel poses.

\section{Implementation Details}\label{sec:3}

In this section, we include implementation details of our proposed method (see also~\cref{tab:hyper}). Our implementation is based on PyTorch~\cite{paszke2019pytorch}. We built our architecture upon 3DGS-Avatar~\cite{qian20233dgs}, but with some important modifications in order to learn multiple identities (see also Sec.~3.2 of the main paper). We use $N_g = 5\times 10^4$ 3D Gaussians, that we initialize by randomly sampling $N_g$ points on the canonical SMPL mesh surface of the first identity. We initialize the matrices $\bm{U}_1$, $\bm{U}_2$, $\bm{U}_3$ as described in Sec.~4.3 of the main paper. 

The positions $\bm{\mu}_c$ of the Gaussians in the canonical space are first encoded into a multi-level hash grid and passed through the non-rigid deformation MLP $f_d$. The hash grid has 16 levels of 2-dimensional features each, hash table size $2^{16}$, coarse and fine resolution 16 and 2048 correspondingly~\cite{qian20233dgs}. The non-rigid MLP $f_d$ is also conditioned on a latent code $\bm{z}_p$ that is the output of an hierarchical encoder~\cite{LEAP:CVPR:21,qian20233dgs}. The rigid MLP $f_r$ inputs the non-rigidly deformed positions $\bm{\mu}_d$ and outputs the skinning weights that sum up to 1 through a softmax layer~\cite{qian20233dgs}. Similarly with 3DGS-Avatar~\cite{qian20233dgs}, we normalize the coordinates in the canonical space by proportionally padding the bounding box enclosing the canonical SMPL mesh of the identity. The color MLP is conditioned on the output $\bm{z}$ of the non-rigid network, a per-Gaussian feature vector $\bm{f}$ and the canonicalized viewing direction (see Sec.~3.2). We do not learn any per-frame latent codes, in order to avoid any overfitting to the training frames. We use a larger color MLP compared to 3DGS-Avatar, of 3 layers and 256 hidden units each, in order to learn the diverse colors of multiple identities.

We observed that our network is very sensitive to the learning rate of the different components and the initialization, similarly with other 3DGS methods~\cite{kerbl20233d}. Our initialization is described in Sec.~4.3. We experimentally chose the learning rates depicted in~\cref{tab:hyper}. As shown, we use different learning rates for different rows of $\bm{U}_1$ that roughly correspond to scaling, rotation, features for color, and opacity (see our tensor construction in Sec.~4.1). These learning rates are similar with the learning rates used for scaling, rotation, color, and opacity by other 3DGS implementations~\cite{kerbl20233d,qian20233dgs}. However, in our case, we get the corresponding values after multiplying the matrices $\bm{U}_1$, $\bm{U}_2$, $\bm{U}_3$ using Eq.~(8). We use Adam optimizer~\cite{kingma2014adam}. The rest of the Adam hyper-parameters are set at their default values ($\beta_1 = 0.9, \beta_2 = 0.999, \epsilon = 10^{-8}$).

We train our multi-identity network for $5 \times 10^4$ iterations for 15 subjects, and $10^5$ iterations for 30 subjects, that need about 2 and 4 hours correspondingly on a single NVIDIA Quadro RTX 6000 GPU.
Following prior work~\cite{weng2022humannerf,qian20233dgs}, we freeze everything in the first $10^3$ iterations and train only the rigid MLP. In this way, we better initialize the rigid mapping from the canonical to the observation space based on estimated SMPL parameters, and avoid any noisy gradients in the beginning. After the first $10^3$ iterations, we enable optimization to all the components, except for the non-rigid network that is frozen until the first $3 \times 10^3$ iterations. In constrast to 3DGS-Avatar~\cite{qian20233dgs}, we do not add any learnable module for pose correction for the estimated SMPL parameters, in order to avoid overfitting to the training body poses.

We use the same loss function as 3DGS-Avatar~\cite{qian20233dgs}:
\begin{equation}\label{eq:tensor_w}
\begin{split}
    \mathcal{L} = \lambda_{l1} \mathcal{L}_{l1} + \lambda_{perc} \mathcal{L}_{perc} + \lambda_{mask} \mathcal{L}_{mask} + \lambda_{skin} \mathcal{L}_{skin} + \\
   \lambda_{isopos} \mathcal{L}_{isopos} + \lambda_{isocov} \mathcal{L}_{isocov}   \;,
\end{split}
\end{equation}
where $\mathcal{L}_{l1}$ is the L1 photometric loss and $\mathcal{L}_{perc}$ is the perceptual (LPIPS) loss with VGG as backbone. The mask loss $\mathcal{L}_{mask}$ corresponds to the L1 loss between the ground truth foreground mask and the predicted mask by accumulating the predicted opacities~\cite{qian20233dgs}. The skinning loss provides a regularization on the non-rigid MLP~\cite{qian20233dgs}. The as-isometric-as possible losses $\mathcal{L}_{isopos}$ and $\mathcal{L}_{isocov}$ restrict the 3D Gaussian positions to preserve a similar distance after deformation to the observed space, with also similar covariance matrices~\cite{qian20233dgs}.
We set $\lambda_{l1} = 1$, $\lambda_{perc} = 0.01$, $\lambda_{mask} = 0.1$, $\lambda_{isopos} = 1$, $\lambda_{isocov} = 100$, and $\lambda_{skin} = 10$ for the first $10^3$ iterations that is then decreased to $\lambda_{skin} = 0.1$~\cite{qian20233dgs}.

\begin{figure}[tb]
  \centering
    \includegraphics[width=\linewidth]{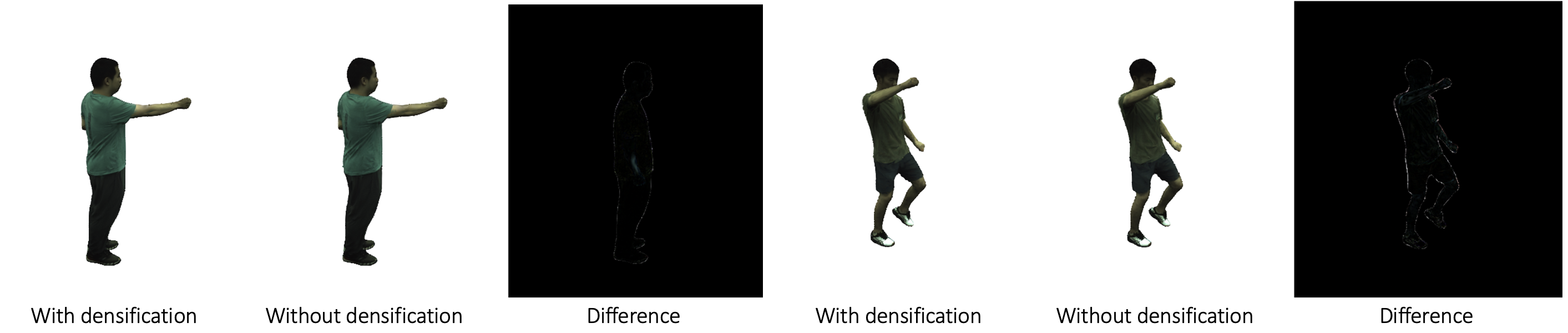}
  \caption{With or without adaptive densification scheme for the 3D Gaussians~\cite{kerbl20233d}.}
  \label{fig:nodense}
\end{figure}

\begin{figure}[tb]
  \centering
  \includegraphics[width=\linewidth]{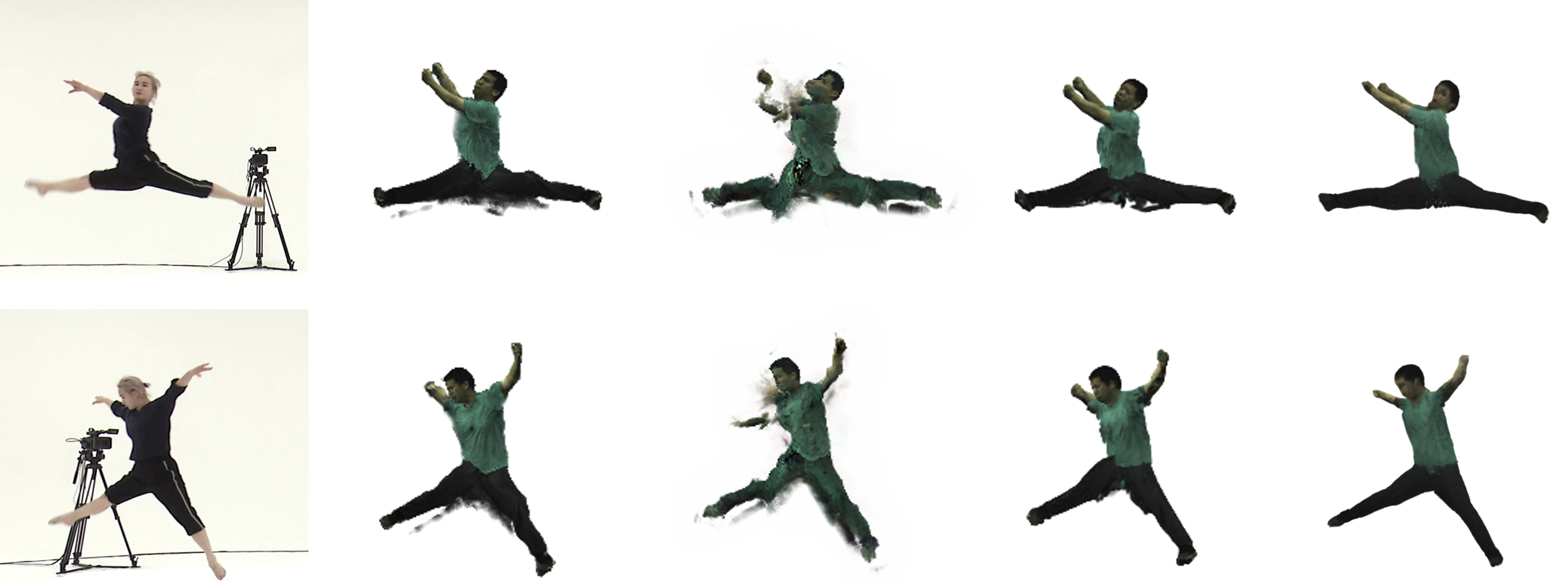}
\includegraphics[width=\linewidth]{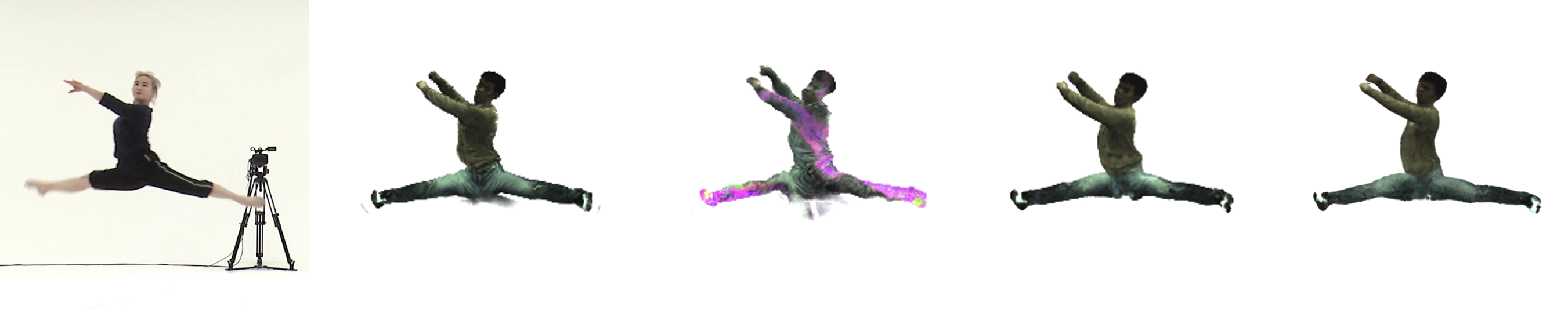}
\includegraphics[width=\linewidth]{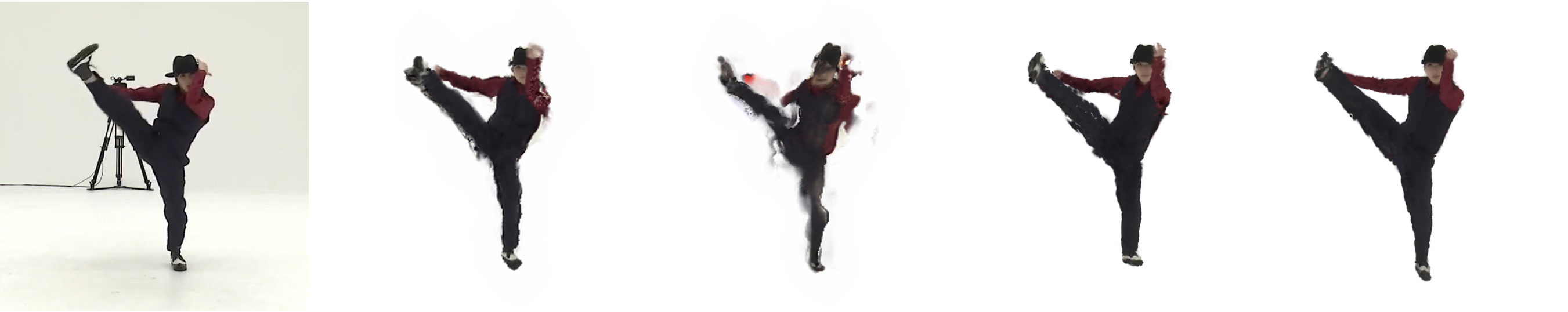}
\includegraphics[width=\linewidth]{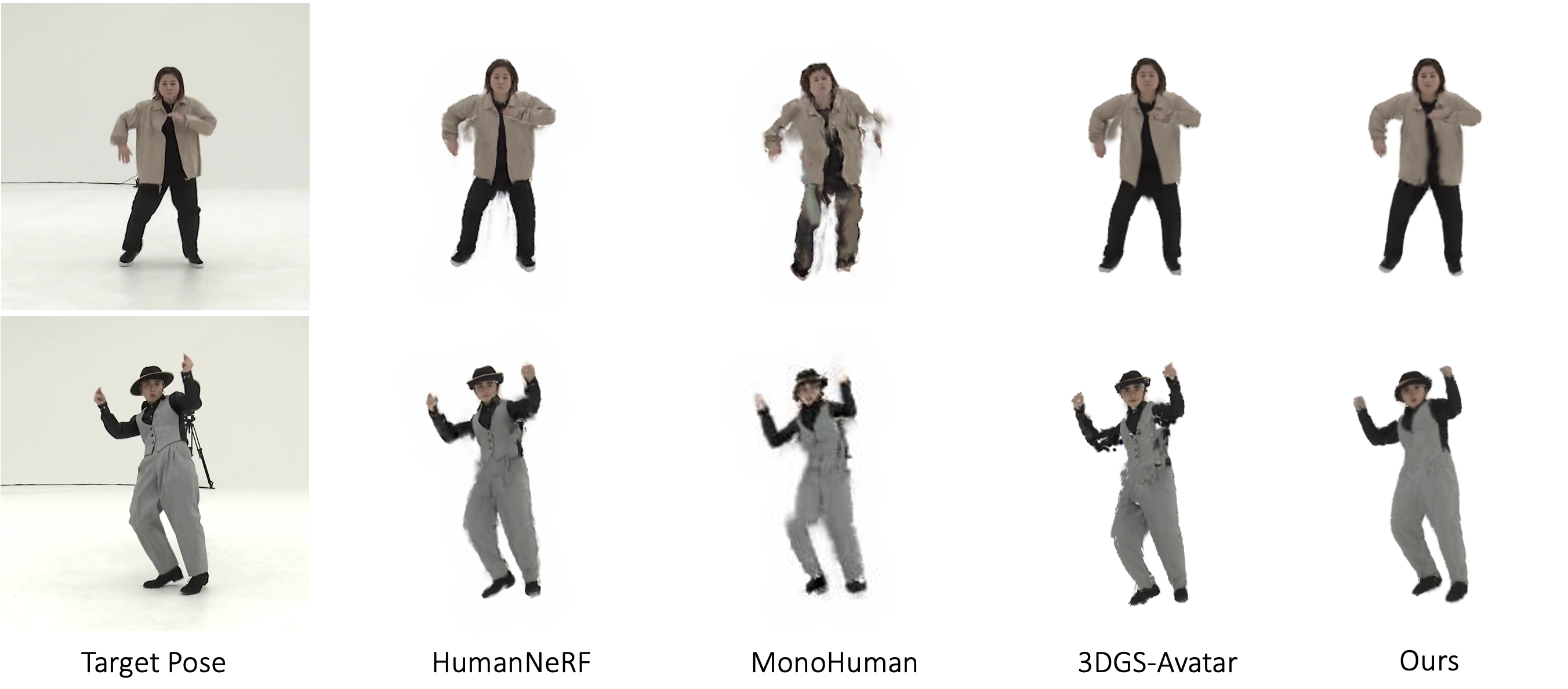}
  \caption{
  \textbf{Animation of human avatars under novel poses.} 
  Qualitative comparison with state-of-the-art approaches, namely HumanNeRF~\cite{weng2022humannerf}, MonoHuman~\cite{yu2023monohuman}, and 3DGS-Avatar~\cite{qian20233dgs}. The subjects are from ZJU-MoCap~\cite{peng2021neural} and AIST++~\cite{li2021learn,aist-dance-db} datasets. The target poses (column 1) are unseen during training, from unseen camera views and advanced dance videos.
  Our method robustly animates all the identities under challenging novel poses, outperforming the other methods.
  }
  \label{fig:pose_aist_suppl}
  \vspace{-15pt}
\end{figure}

In order to avoid any non-differentiable gradient updates, we do not apply any densification or pruning of the 3D Gaussians during training, in contrast to the original 3DGS implementation~\cite{kerbl20233d}. Instead, we keep $N_g$ Gaussians throughout our optimization, which are moved and deformed according to our network. In this way, the learnable matrices $\bm{U}_1$, $\bm{U}_2$, $\bm{U}_3$ are directly optimized with gradient descent, without any non-differentiable updates, and include all parameters for all the training identities. In our preliminary experiments, we observed that we achieve similar results for a single identity with or without densification and pruning, using $N_g = 5 \times 10^4$ (see \cref{fig:nodense}). However, including an adaptive density control of the Gaussians can be explored as future work.

For $N_i = 30$ identities, MIGS learns only $(M+N_i+N_g) R = (43 + 30 + 5\times 10^4)\times 100 \approx 5 \times 10 ^ 6 $ parameters, compared to $M N_i N_g \approx 6.5 \times 10^7$ that would be required by single-identity 3DGS representations, leading to a decrease \emph{by at least one order of magnitude} in the total number of learnable parameters.  


\begin{table}[t]
\begin{center}
\scalebox{0.85}{
\begin{tabular}{|l|c|}
\hline
Number of 3D Gaussians $N_g$ & $5 \times 10^{4}$ \\
Rank of CP tensor decomposition $R$ & $100$ \\
Dimension of feature vector $\bm{f}$ & $32$ \\
Dimension of non-rigid output vector $\bm{z}$ & $32$ \\
Non-rigid deformation MLP $f_d$: Linear layers & 3 \\
Non-rigid deformation MLP $f_d$: Hidden units & 128 \\
Rigid transformation MLP $f_r$: Linear layers & 3 \\
Rigid transformation MLP $f_r$: Hidden units & 128 \\
Color MLP $f_c$: Linear layers & 3 \\
Color MLP $f_c$: Hidden units & 256 \\
Activation & ReLU \\
Optimizer & Adam \\
Learning rate for $\bm{U}_{1_{3:6,:}}$ (scaling) & $5 \times 10^{-3}$ \\
Learning rate for $\bm{U}_{1_{6:10,:}}$ (rotation) & $10^{-3}$ \\
Learning rate for $\bm{U}_{1_{10:42,:}}$ (feature) & $2.5 \times 10^{-3}$ \\
Learning rate for $\bm{U}_{1_{42:43,:}}$ (opacity) & $5 \times 10^{-2}$ \\
Initial learning rate for $\bm{U}_{1_{:3,:}}$, $\bm{U}_{2}$, $\bm{U}_{3}$ & $1.6\times10^{-4}$\\
Final learning rate or $\bm{U}_{1_{:3,:}}$, $\bm{U}_{2}$, $\bm{U}_{3}$ & $1.6\times10^{-6}$\\
Initial learning rate for rigid MLP & $10^{-4}$ \\
Initial learning rate for non-rigid and color MLP & $10^{-3}$ \\
Final learning rate for MLPs & $10^{-6}$ \\
Learning rate schedule & exponential decay \\
Max iterations & $10 ^ 5$ \\
\hline
\end{tabular}}
\end{center}
\caption{Hyper-parameters of our architecture.}
\label{tab:hyper}
\end{table}

\section{Limitations}\label{sec:5}

We observe that in some cases, our multi-identity network may fail to capture fine-grained details, such as high-frequency texture in clothes or facial details. This tends to happen more when the number of identities increases beyond 20, since the network leverages information from multiple identities for learning, thus smoothing the result. In our work, we tackle this smoothing with the personalization procedure (see Sec.~4.4 and \cref{fig:ablation_personal}). In the future, we plan to further enhance high-frequency details by using higher-resolution data and exploring other tensor structures. In addition, in the technical component, we have no theoretical proof that the CP decomposition is the optimal way to factorize the parameters, but we notice empirically that this suffices in our particular case. 


\section{Ethical Considerations}\label{sec:6}

We would like to note the potential misuse of video synthesis methods. With the advances in neural rendering, recent methods can generate photo-realistic human avatars. Our research focuses on human body animation and we presented as main application the animation of human avatars under challenging dance sequences. In contrast to deep fakes, we believe this application cannot be used to spread misinformation or for other harmful purposes. However, we would like to emphasize that there is still a risk of using such methods to generate misleading content. Thus, research on fake content detection and forensics is crucial. We intend to share our source code to help improving such research.

\end{document}